\documentclass[journal]{IEEEtran}
\ifCLASSINFOpdf
\else
\fi
\usepackage{url}

\usepackage{graphicx}
\usepackage[linesnumbered, ruled,vlined]{algorithm2e}

\usepackage{amssymb}
\usepackage{amsmath}

\usepackage{subfig}
\usepackage{float}
\usepackage{ifthen}
\usepackage{algorithmic}

\newlength{\commentWidth}
\begin{document}
\bstctlcite{IEEEexample:BSTcontrol}

\title{Transfer Dynamics in Emergent Evolutionary Curricula}


\author{
    \IEEEauthorblockN{Aaron Dharna\IEEEauthorrefmark{1}, Amy K Hoover\IEEEauthorrefmark{1}, Julian Togelius\IEEEauthorrefmark{2}, L. B. Soros\IEEEauthorrefmark{3}}\\\
    \IEEEauthorblockA{\IEEEauthorrefmark{1}New Jersey Institute of Technology, USA, aadharna@gmail.com}\\\
    \IEEEauthorblockA{\IEEEauthorrefmark{2}New York University, USA}\\\
    \IEEEauthorblockA{\IEEEauthorrefmark{3}Cross Labs, Cross Compass Ltd., Tokyo, Japan
    }
}

\markboth{}%
{Shell \MakeLowercase{\textit{et al.}}: Bare Demo of IEEEtran.cls for IEEE Journals}

\maketitle

\begin{abstract}
 
PINSKY is a system for open-ended learning through neuroevolution in game-based domains. It builds on the Paired Open-Ended Trailblazer (POET) system, which originally explored learning and environment generation for bipedal walkers, and adapts it to games in the General Video Game AI (GVGAI) system. Previous work showed that by co-evolving levels and neural network policies, levels could be found for which successful policies could not be created via optimization alone. 
Studied in the realm of Artificial Life as a potentially open-ended alternative to gradient-based fitness, minimal criteria (MC)-based selection helps foster diversity in evolutionary populations.
The main question addressed by this paper is how the open-ended learning actually works, focusing in particular on the role of transfer of policies from one evolutionary branch (``species'') to another. We analyze the dynamics of the system through creating phylogenetic trees, analyzing evolutionary trajectories of policies, and temporally breaking down transfers according to species type. Furthermore, we analyze the impact of the minimal criterion on generated level diversity and inter-species transfer. The most insightful finding is that inter-species transfer, while rare, is crucial to the system's success. 
\end{abstract}


%
\IEEEpeerreviewmaketitle

\section{Introduction}

\IEEEPARstart{O}{pen}-ended learning is a long-standing goal of AI research, and is arguably one of the more promising approaches toward achieving more general artificial intelligence. The goal in open-ended learning is to make agents learn not just to perform a given task, but to continually learn a complex repertoire that is appropriate across a growing set of situations. 
In contrast, much current reinforcement learning (RL) and evolutionary computation research focuses on learning narrow solutions to specific problems, which risks significant overfitting and requires human designers to specify problems and rewards, limiting the long-term learning potential.

Paired Open-Ended Trailblazer (POET) \cite{wang:gecco19} is a recently-proposed method for open-ended learning. This method keeps a meta-population of agent-environment pairs.
Agents are trained to perform well on their environments, and new environments are evolved so as to challenge the agents. POET is noteworthy because it generates emergent curricula of diverse biped walking challenges from scratch, i.e. with no human input. 
Intermittently, agents are tested on other generated environments than their own, and can transfer there (replacing the agent that was originally paired with the environment). Experiments in \cite{dharna2020cogeneration} and \cite{wang2020enhanced} showed that standard RL and evolutionary computation optimization algorithms were unable to be trained to solve complex POET-generated environments without this agent transfer mechanism. 

The setting used for the original POET environment was at the same time quite limiting and lacking certain complications typical of learning in other settings. In particular, the dense reward function for the walker agents makes it easy to learn good behaviors, and most walker environments that can be generated are also solvable by agents. Therefore PINSKY (introduced below) adapts POET to a more complex domain and reward-sparse setting, which would enable harder problems to be created and which allow for more diverse environments.

The new system utilizes the General Video Game AI (GVGAI) system \cite{perez2014gvgai}, and in particular uses variants of GVGAI's Zelda game, namely Deterministic Zelda (dZelda). Zelda, as used here, is a 2D game about maze navigation and combat. In order to make the POET loop work, several modifications (described later in this paper) had to be made. Importantly, agents based on Monte Carlo Tree Search (MCTS) \cite{MCTS2006} were needed to verify the solvability of environments, as environments could be generated that would not be solvable by \emph{any} agents, in striking contrast to the walker environment.

We call this new system PINSKY -- POET-Inspired Neuroevolutionary System for Kreativity.\footnote{Code: \url{https://tinyurl.com/y42x54pg}} A previous conference paper \cite{dharna2020cogeneration}, which this paper is an extension of and follow-up to, demonstrates that basic functioning of the PINSKY system. This paper investigates \emph{how} PINSKY works. Generated levels are speciated into distinct clusters, and then agent transfer dynamics are analyzed over multiple runs, focusing on transfer between species. Furthermore, this paper investigates what happens when the MCTS-based solvability test is removed. Here, the hypothesis is that not checking for solvability would a) allow unsolvable environments to swamp the system and b) collapse the generated level diversity thereby inhibiting inter-species transfer and eventually grinding progress to a halt. 

Understanding POET's transfer mechanism in an evolutionary context may provide insight into transfer mechanisms in gradient-based machine learning. Do individuals tend to transfer into very unfamiliar environments, or only those that are similar to the environment on which a given agent has already proven successful? Do individuals only transfer to tasks of increasing complexity, or can they transfer back and forth between simple and complex tasks in order to find agents that solve difficult tasks? Answering these questions will likely bring us closer to the goal of creating agents that are capable of applying learned skills in multiple contexts. To this end, the main contribution of this paper is an analytical framework for studying transfer dynamics in POET-like systems.

\section{Related Work}


This section provides an introduction to transfer and curriculum learning,  adversarial training, and coevolution before introducing the POET algorithm as a coevolutionary algorithm. POET-style coevolutionary algorithms provide a unique setting to study the interactions of transfer learning with optimization and coevolutionary dynamics on the ability to discover curricula and high-functioning behaviors.

\subsection{Transfer and Curriculum Learning}


A popular paradigm in machine learning is transfer learning, which rather than training from a randomly initialized model on the target task, instead first pre-trains the model on one or more source tasks. The idea is that shared problem structure between the source and target tasks is captured in what the model learns during training on the source tasks (e.g., captured in artificial neural networks' (ANN) weights or Gaussian mixture models' parameters \cite{gmmTransfer2019}). Transfer learning exploits this shared structure to increase performance or decrease training time on the \emph{target} task \cite{bengioCL2009}\cite{Lazaric2008batch}\cite{Taylor09Negative}\cite{wangTranfserGP2020}\cite{karpathy2014deep}\cite{finn2017modelagnostic}. The hope is that the pre-training search identifies a point on the surface of the optimization landscape that is close to points of high performance with respect to the target task \cite{finn2017modelagnostic}\cite{shelhamer2016loss}\cite{alex2019evolvability}.

While sub-fields of machine learning describe versions of transfer learning particular to the paradigm, the base components are a task $\psi_i$ and model $\theta_i$. Transfer learning for control problems often defines the task $\psi_i$ as a MDP $M_i$. This MDP is defined by the tuple of functions that parameterize States, Actions, Transitions, Rewards, and the discount factor ($\mathcal{S}_i$, $\mathcal{A}_i$, $\mathcal{T}_i$, $\mathcal{R}_i$, $\gamma_i$) \cite{narvekarCurriculum}. Typically an MDP $M_i$ is solved when a policy, $\pi_{\theta_i}(a|s)$ -- a learnable model with parameterization $\theta_i$ mapping states to actions -- maximizes the long-term discounted sum of rewards $\sum_{k=t+1}^T \gamma^{k-t-1} r_k$ \cite{MDP}. An example of fine-tuning to task $\psi_j$ from task $\psi_i$ is solving $M_j$ with $\pi_{\theta_i}$. $\psi_i$ and $\psi_j$ can differ in one or more of the five components of an MDP.

Due to the larger definition of an MDP as a task, many different types of knowledge can be learned and transferred between tasks.
When used in RL, Q-functions (often but not always represented as ANNs) provide information on how valuable an action is in a given state and can be transferred to untrained RL agents to speed up agent training \cite{shelhamer2016loss}. Transferring knowledge (stored in DQNs) between various scenarios in 3D environments has proven helpful for generalizing agents to traverse unknown maps with unknown backgrounds in the VizDoom environment \cite{vizDoom2016}. If the laws that govern a system (e.g. physical laws like gravity) are consistent, then a learned dynamics model would conserve such knowledge even if objectives were to change \cite{nagab2018learning}.  A policy indicates which actions are potentially useful, therefore one might transplant effective policies between tasks \cite{finn2017modelagnostic}\cite{Vinyals2019AStar}\cite{song2020rapidly}. Alternatively, transfer learning has been used in multi-agent settings to transfer policies between multiple agents in a shared simulation, e.g. StarCraft II \cite{StarCraftTransfer2018}. Transferring a policy into a new task/environment for additional fine-tuning is POET's form of transfer learning; after POET creates a new agent-environment pair (thereby successively altering $\mathcal{S}_i$ in the definition of an MDP), the agent will receive additional inner-loop optimization to adapt its policy to the new environment from the starting point of the parent's behavioral policy. 


Continually complexifying various aspects of an initial simple MDP, thereby creating a regimen of tasks for the purpose of scaffolding agent skill acquisition, is known as curriculum learning. Specifically, Narvekar et al. define curriculum learning (in RL) as learning a directed acyclic graph of a set of MDPs \cite{narvekarCurriculum} where the goal is to find the correct sequencing of experience/tasks such that policy $\pi_{\theta_i}(a|s)$ can solve difficult tasks. This set of MDPs is often defined a priori, or, as in POET, generated on the fly. In POET, the generated MDP coevolves with respect to the agent performance.



\subsection{Adversarial Training, Self-Play, and Coevolution}

Called generative adversarial training by Goodfellow et al. \cite{GoodfellowGAN} and self-play in modern approaches in reinforcement learning \cite{Vinyals2019AStar}\cite{Arulkumaran_2019}\cite{SilverGo2016}\cite{openai2019dota}, cooperative and competitive co-evolution are at the heart of many high-profile algorithmic achievements \cite{chellapilla:tec01}. Unlike traditional evolutionary systems, in coevolutionary settings individuals are compared based on the outcome of their interaction with other members of the population(s) rather than only comparing each individual w.r.t. a fixed fitness function. As a result, coevolutionary search is a stochastic or non-stationary process \cite{Popovici2012}.

Coevolution is also at the heart of an approach for discovering tree-based policies and constructing feature extractors that with minimal model complexity compete with deep-learning based solutions to play the 49 games in the Arcade Learning Environment (ALE) \cite{Kelly2018CoevoAtari}. Beyond simply highly skilled solutions, Kelly and Heywood discovered reusable subtrees across multiple games inside the coevolved policies \cite{Kelly2018CoevoAtari}. Blondie 24, a single-population coevolutionary setup, taught itself to play checkers competitively by evolving the weights of a fixed-architecture ANN and evaluating performance based on how well each agent plays against a set of opponents drawn from the current population \cite{chellapilla:tec01}. 

In two-player competitive games, self-play 
and coevolution intersect when policies compete against each other to supply data that informs the future trajectory of search \cite{Lanctot2017Unified}\cite{Popovici2012}. In a GAN the game-theoretic approach of having a generator and discriminator that compete against each other in a zero-sum game shows one scenario where the gradient information is determined by interactions between a competitive population of two \cite{guttenberg2018potential}\cite{Popovici2012}. The challenge for solving problems with self-play or coevolution lies in finding the Nash equilibrium such that the distributions of each population are equally aligned. However, the subjective nature of the competition between members of the population(s) can lead to instability as search in coevolutionary systems is dependent on interactions between individuals in the population \cite{guttenberg2018potential}.

\section{POET and PINSKY}

\subsection{Paired Open-Ended Trailblazer (POET)}
The POET algorithm \cite{wang:gecco19}\cite{wang2020enhanced} is a coevolutionary algorithm for simultaneously generating and solving tasks.
The approach was first explored on the OpenAI Gym's Hardcore Bipedal Walker domain;\footnote{\url{https://gym.openai.com/envs/BipedalWalkerHardcore-v2/}} given rangefinders and joint angle information, agents learn gaits for walking over terrain containing hills, stumps, and pitfalls. 
POET coevolves agents and terrains through three main processes: 1) periodically generating new offspring environments by mutating existing environments and ensuring that new environments pass a \emph{minimal criterion} (MC) for reproduction of being neither too easy nor too hard, 2) incrementally optimizing agents paired with generated environments, and 3) occasionally attempting to transfer optimized agents into other environments in the meta-population (Algorithm \ref{alg:gvgaiPOET}). POET-style algorithms eventually create difficult environment instances that cannot be solved by optimization from scratch (starting from a random policy), indicating that transfer is necessary for high-performing behaviors \cite{wang:gecco19}\cite{wang2020enhanced}\cite{dharna2020cogeneration}.






\begin{figure*}[ht]
  \includegraphics[width=\textwidth]{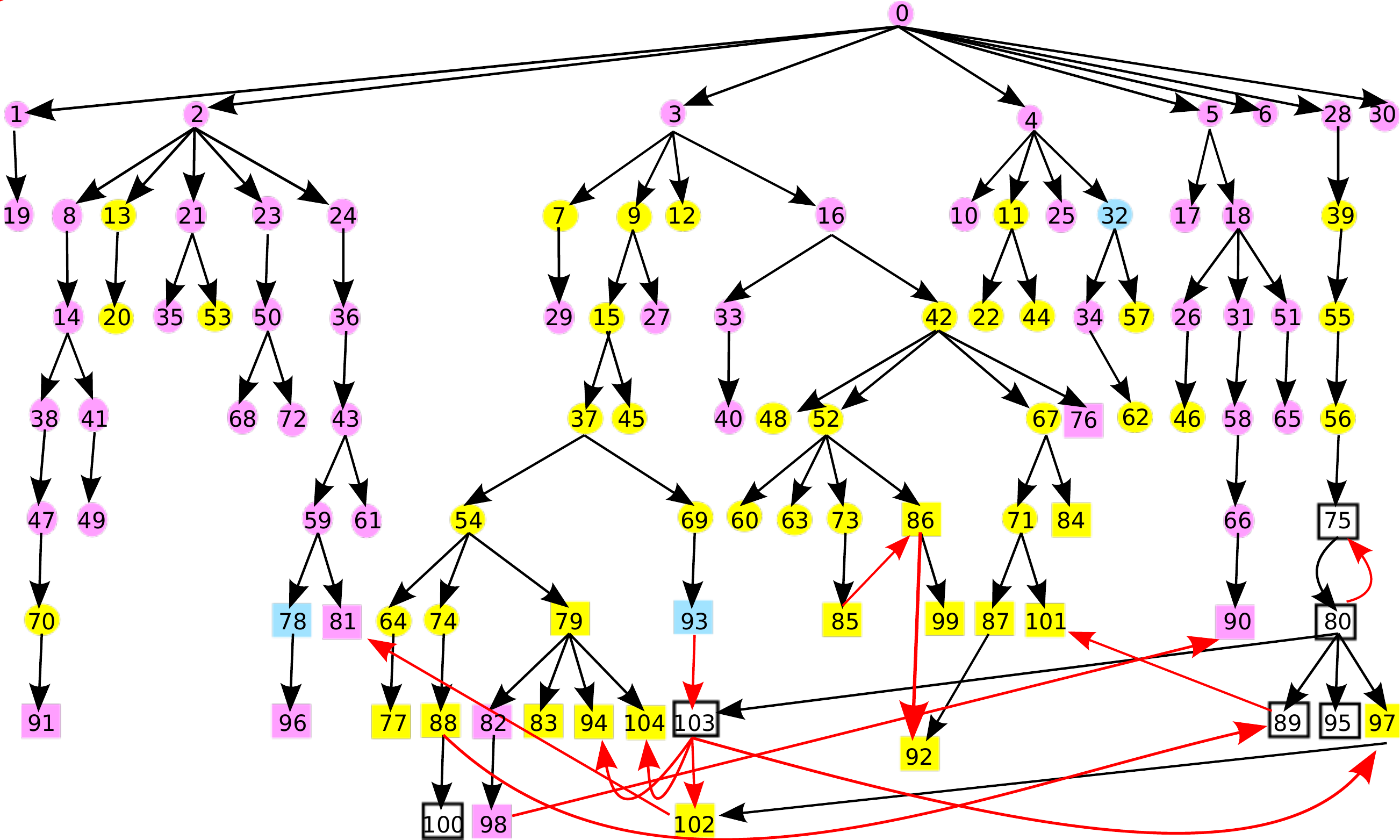}
  \caption{\textbf{Phylogenetic tree of the first 105 environments created by PINSKY in the first multiDoor dZelda experiment.} Black arrows denote lineage showing children environment IDs coming from parent environment IDs. Red arrows show \textbf{one} step of transfer between the current agent-environment population in PINSKY. All transfer happens concurrently. Color of the node denotes the species that the environment belongs to. Square vertices (75 to 104) are in PINSKY's active population.} 
  \label{fig:manyDoorsTree}
\end{figure*}

The reward function for biped walkers is continuous, allowing ``neither too easy nor to hard'' (with respect to the MC) to be defined by minimal and maximal acceptable reward values. This binary approach to fitness, explored recently in the context of artificial life and evolutionary robotics \cite{LehmanMC}, presents a potentially more open-ended alternative to
traditional gradient-based evolution \cite{Soros2014IdentifyingNC}\cite{soros18Thesis}. After an environment satisfies the difficulty criteria, the new environment is paired with a copy of its parent's paired agent. Another important and unusual feature of POET is that it periodically evaluates all possible pairs of agents and environments in the meta-population, thereby
revealing behaviors that can be easily adapted to multiple environments. Between transfer attempts, agents optimize exclusively on their paired source environment. The best candidate agent for transfer into a level is determined via tournament selection (Algorithm \ref{alg:transfer}) based on each agent's zero-shot performance on each environment where ties break in the favor of the incumbent agent. Through incremental optimization and regular transfer of agents, POET generates curricula and agents for biped walking.



POET's evolutionary framing of environment generation inherently induces the necessary directed acyclic relationship between the initial specified level and all future levels (see lineage structure in Figure \ref{fig:manyDoorsTree}) that is required to meet the definition of curriculum learning supplied by Narvekar et al. \cite{narvekarCurriculum}. However, the automatic generation of new tasks via an evolutionary generative process allows the curriculum to be built by POET itself in an online manner that is reactive to the minimal criterion and agent population.

\begin{algorithm}[t]
\DontPrintSemicolon
\SetAlgoLined
 Pair initial environment with unoptimized agent\\
 \While(\tcp*[h]{Outer Loop}){not done}{
  \If{ counter \% mutationTimer == 0}{
    Generate offspring environment-agent pairs\\ 
    Remove too-easy and too-difficult offspring \\
    \If{population size exceeded}{Remove oldest environment-agent pairs}
  }
  Optimize for a batch of steps \tcp*[h]{Inner Loop}\\ 
  Reevaluate all optimized individuals 
  
  \If{counter \% transferTimer == 0}{
    Evaluate all agents on all environments \\
    Replace incumbent agents with more successful agents, if any exist
  }
  counter += 1
 }
 \caption{POET-style Algorithms}
 \label{alg:gvgaiPOET}
\end{algorithm}

\begin{algorithm}[t]
  \DontPrintSemicolon
  \SetAlgoLined
    \SetKwInOut{Input}{input}

    \Input{t: POET loop id. \newline 
           P: Current meta-population of agent-env pairs.}

    E = Evaluate $p_j$.agent in $p_i$.environment $\forall p_i \text{, } p_j \in$ P.\\
    \tcp{$E_{ij}$ = score of $p_j$.agent in $p_i$.environment}
    \For{$p_i \in$ P}{
        \tcp{Find pair $\in P$ containing the best agent for environment $i$}
        $p_{\text{best}}$ = argmax($E_i$)\\
        \If{not $p_i$ == $p_{\text{best}}$}{
            Update $p_i$.agent to $p_{\text{best}}$.agent\\
        }
    }
    \caption{Tournament Update: transfer mechanism. Algorithm 2 expands lines 13 and 14 of Algorithm 1.}
    \label{alg:transfer}
\end{algorithm}

\subsection{PINSKY and Differences from POET}

The POET-Inspired Neuroevolutionary System for KreativitY (PINSKY) is an adaptation of the POET algorithm from an evolutionary robotics domain into the space of 2D Atari-style games in the General Video Game Artificial Intelligence (GVGAI) framework \cite{perez2014gvgai}\cite{perezliebana2018general}. GVGAI provides an interface for defining and playing games written in Video Game Description Language (VGDL) \cite{schaul2014extensible}, a human-readable text language for 2D games and levels ranging from dungeon crawlers to platformers. Adapting POET to this new set of game environments required a number of changes to the underlying POET algorithm. 

\subsubsection{Reward Function}

In POET, the reward function is both dense (making it easy to tease out subtle differences between agents) and well-aligned to the desired task (making the optimization process continuously steer the agent towards improvement). In game environments, neither of these conditions are necessarily met.

The RL problem of credit assignment, or determining which actions cause the observed outcome, is hard even when the reward function is dense. A sparse reward function further complicates this task making games, such as the GVGAI games Solarfox and Deterministic Zelda (dZelda), substantially more difficult than the biped walker domain. The dZelda game functions similarly to Zelda (a dungeon crawler where the agent needs to avoid/kill enemies and escape by getting a key to the door), but the enemies greedily chase the avatar rather than move randomly. POET uses the OpenAI ES \cite{salimans2017:ESRL} algorithm to optimize agent's long-term reward. PINSKY parameterizes the inner loop optimization algorithm, and by default uses Differential Evolution \cite{StornDE} to optimize a real-valued vector of the policy weights; however, any optimization algorithm can serve this function.\footnote{Preliminary experiments explored PINSKY combined with PPO, OpenAI's ES, CMA-ES and other optimization algorithms in the inner loop.} 

The dZelda agent is rewarded for picking up a  key, taking it to the door (the win condition), and killing monsters. However, the additional reward source of killing monsters makes the reward function unaligned to the desired task; solely killing monsters can earn more reward than winning the game, providing a distracting reward. Furthermore, the reward can be sparse, making the task of differentiating similar agents increasingly difficult. For example, an agent that gets two points for killing a monster and picking up a key is the same in terms of reward as an agent that only kills two monsters; therefore, the reward signal is not as clear as the biped walker domain's reward signal that was explored in POET. 

\subsubsection{Minimum Playability Criteria}

POET prevents evolutionary search from degenerating by requiring that evolved terrains satisfy a minimal criterion \cite{LehmanMC} defined \textit{a priori}; the walker has to be able to walk at least a minimum amount (ensuring the level is not too hard) and at most a maximum amount (ensuring the level is not too easy). In PINSKY, the minimal criterion concept has been adapted into a playability criterion due to the sparsity of reward in games. Specifically, a level is too easy if a random agent \textit{can} beat the level and too hard if an MCTS agent (with the default GVGAI time limit of 40ms of planning time per action) \textit{cannot} beat the level. Methods such as MCTS are limiting because having a fast forward model is often an onerous requirement. Furthermore, even with a fast forward model, planning algorithms like MCTS are still subject to variable performance \cite{Nelson2016MCTS}. Nevertheless,  MCTS is robust enough to function as a simple playability check that can solve a variety of complex levels for these particular games. 

The MC combined with age-based culling allows evolutionary drift to introduce new challenges (represented internally as a direct encoding of tile positions) that the neural network agents will adapt to. The random mutation in the MC-based evolutionary level generator can add, remove, or move tiles and is slightly biased towards adding new game objects (enemies, walls, keys, and optionally additional doors) into the levels, disrupting existing policies. Meanwhile, culling by age provides ample time for the entire population of agents to attempt to solve the new task through direct optimization of the paired agent and repeated transfer attempts of all other agents to replace the paired agent.




\begin{figure}[bp!]
  \centering
  \includegraphics[width=0.45\textwidth]{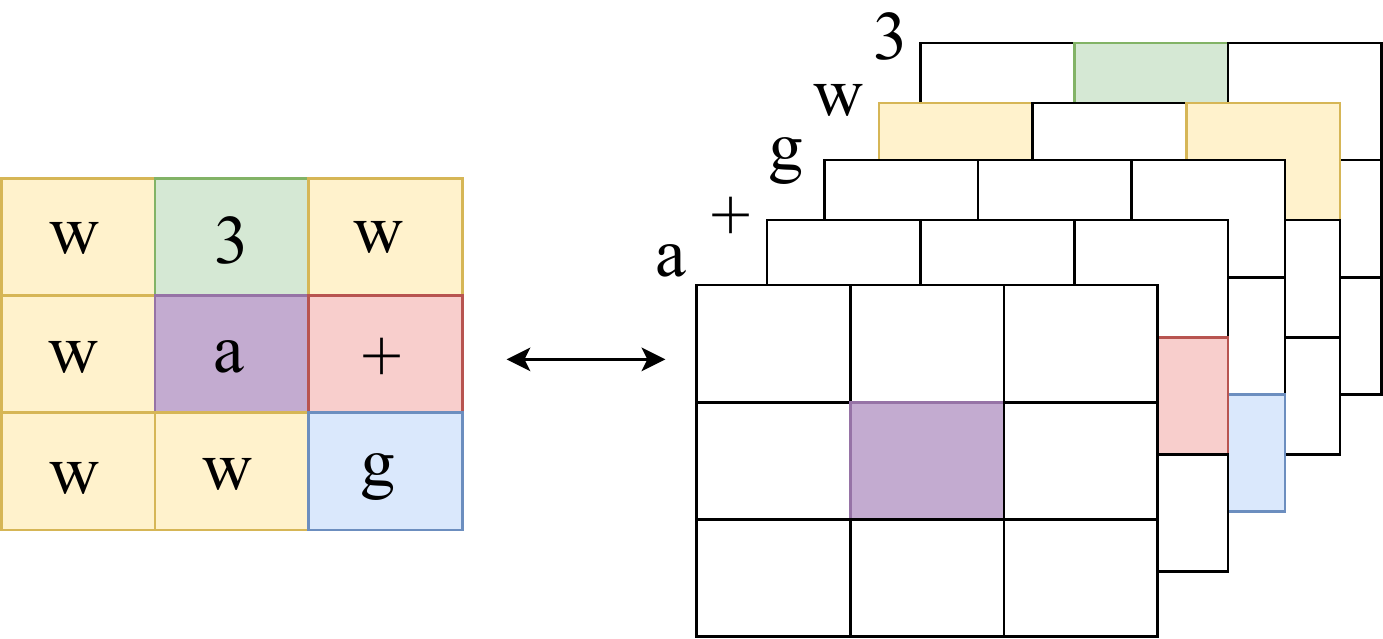}
  \caption{\textbf{One-hot encoded map input to the convolutional policy network.} Tiles in each GVGAI map (left) correspond to $x$,$y$ positions in the environment. In this example from dZelda, possible tiles include (w)all, (g)oal, (a)vatar, key (+), and monster (3). These 2D maps are then extended into a tensor (right) where each slice denotes the presence (indicated by color) or absence of each tile type.}
  \label{fig:mapTensor}
\end{figure}

\subsubsection{ANN Input}
\begin{figure}[bp!]
  \centering
  \includegraphics[width=0.45\textwidth]{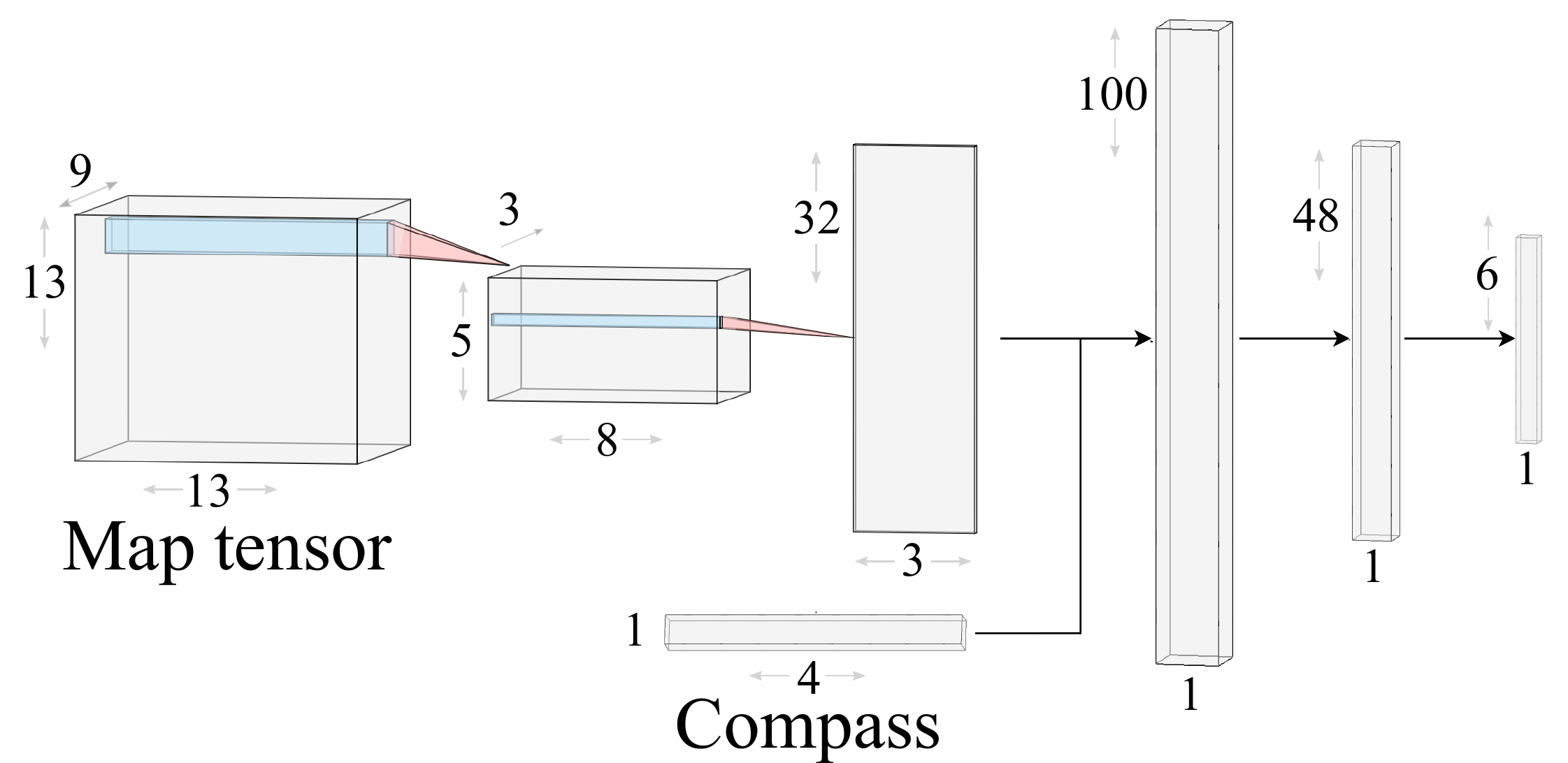}
  \caption{\textbf{Dual-input convolutional policy network for dZelda.} As input, the network takes both the one-hot encoded GVGAI map and agent orientation to produce an action.}
  \label{fig:TileNet}
\end{figure}

The POET agent had access to rangefinder readings and information about its own joint angles. In this agent-centric paradigm, each action results only from local state information.
In contrast, PINSKY agent neural networks are given a fully-observable tile map of the environment and agent orientation (Figures \ref{fig:mapTensor} and \ref{fig:TileNet}). Moving away from purely agent-centric network inputs potentially enables the generalization of PINSKY to arbitrary 2D Atari-style games.

A hyperparameter list controlling various aspects of the nested evolutionary processes of PINSKY's meta-population is given by Table \ref{table:GVGAIPOETarguments}. Our previous experiments \cite{dharna2020cogeneration} and previous experiments by Wang et al. \cite{wang2020enhanced} show that the transfer dynamics in POET-style systems are a necessary component to the system. However, little is known about \emph{how} the transfer dynamics help bootstrap POET-style systems. 

\begin{table}[t]
  \centering
  \begin{tabular}{ |p{2cm}|p{1.0cm}|p{4cm}|  }
    \hline
    Argument & Default & Description\\
      \hline \hline
  game          & dZelda & GVGAI game to play\\
  maxGameLen    & 500    & Max actions per game\\ 
  nGames        & 1500   & Inner-loop evals per opt. step\\  
  popSize       & 50     & Inner-loop population size\\
  mutationTimer & 25     & Outer-loops before mutation step\\
  maxChildren   & 8      & Max offspring per mutation step\\
  mutationRate  & 0.8    & Parent level mutation rate\\
  transferTimer & 10     & Outer-loops before transfer attempt\\
  maxEnvs       & 30     & Agent-env. pair meta-pop. size\\
  numPoetLoops  & 5000   & Max PINSKY outer-loops\\
  alignedReward & False  & Use built-in or aligned reward fn\\
 \hline
    \end{tabular}
  \caption{PINSKY parameters}
  \label{table:GVGAIPOETarguments}
\end{table}

\section{Methodology}

The main contribution of this paper is an analysis of the transfer dynamics in PINSKY. 
In the original POET work, the underlying assumption is that transfer is necessary because it allows the algorithm to reach parts of the search space that otherwise might go unexplored by individuals with a particular skill. That is, these individuals would need to traverse points in the search space known as stepping stones \cite{lehman2011abandoning}  that bridge gaps in the search space. The hypothesis is that without transfer, these individuals would never otherwise reach these stepping stones necessary for solving difficult tasks. In this paper, we revisit this assumption by analyzing the similarity of generated levels in PINSKY and the transfer dynamics of agents between the levels. 





\subsection{Dataset/Environment} \label{sec:dataset}

Data points are collected from eight runs of PINSKY, which simultaneously generate levels and optimize agent performance. Specifically, PINSKY generates: a genetic lineage of game levels, including all agent-level pairings, the time at which the level was first solved and who first solved it, the time at which an agent transfers into a new environment, and which agent was transferred. The transfer analysis is performed on data from PINSKY runs on dZelda. 
In both Zelda and dZelda, an agent spawns in a specified starting location at the beginning of the game, and completes the level by picking up a key and carrying it to the door. In Zelda, the agent must also avoid/kill monsters that start from specific positions in the level, but move randomly. dZelda removes stochasticity from monster pathing; instead, monsters move greedily toward the agent. The paths of the monsters depend only upon where they originally spawn in the level and the current position of the player agent. 

The traditional reward structure of Zelda (shown below as $R_D$) is incremental, meaning that the agent receives small rewards for its actions during gameplay \cite{perez2014gvgai}. Agents receive a point for picking up a key, a point for taking it to the door, and a point for every monster that it kills during play. 
\[R_D = \begin{cases} 
      +1 & \text{Kills Monster}\\
      +1 & \text{Picks-up Key}\\
      +1 & \text{Opens Door}\\
  \end{cases}
\] 
\noindent However, 
preliminary results show that agents are more concerned with killing monsters than actually winning the game. An alternative reward structure defined as $R_A$, \[R_A = \begin{cases} 
      1 - \frac{n_{steps}}{maxGameLen} & \text{Agent reaches goal} \\
      -1 + \frac{n_{steps}}{maxGameLen} & \text{Agent Dies} \\
  \end{cases}
\]
\noindent focuses on quickly getting to the goal by combining the normalized rollout length with a binary win/loss signal. When the goal is unreachable for a given policy, $R_A$ rewards those agents who live the longest. 
While both schemes reward an agent's incremental progress in a level, $R_A$ also has components of a binary scheme where agents are only rewarded at the end of the game \cite{perez2014gvgai}. The effect for agents that optimize through reinforcement learning is that $R_A$ is sparse with respect to when the agent receives rewards. While a sparser reward scheme often poses greater challenges for most RL algorithms \cite{huang2020action}\cite{pathak2017curiositydriven}, game playing agents in Go \cite{Silver2017} and StarCraft II \cite{Vinyals2019AStar} are equipped with similarly sparse rewards. A scheme such as $R_A$ is generalizable to any game in the GVGAI framework and increased performance in Zelda \cite{dharna2020cogeneration}.

While the primary hallmark of complexification in human-designed game levels is the particular distribution of internal wall segments in relation to the starting position of the agent and the monsters, four of the eight experiments additionally permit construction of multiple doors. In these levels, agents are required to find a key and visit \emph{each} door before completing the level. The data from experiments with this additional constraint are called \emph{multidoor}, while those with the traditional tile constraints are called \emph{singledoor}. The idea is that the extra doors could result in levels that are more difficult to solve. 

Finally, data also contains information from experiments run to determine the impact of the minimal criterion. In POET, viable environments must satisfy a minimal criterion to prevent degeneration. For those experiments, the walker satisfied these conditions if it walked a distance between a minimum and maximum value. In PINSKY, the minimal criterion is interpreted as playability, where a level is too easy if a random agent completes it but too hard if an agent running Monte Carlo Tree Search cannot (where the default time limit per rollout is 40ms). 

Data comes from eight different experiments with a combination of these experimental parameters shown below.

\begin{center}
\begin{tabular}{ c c c c c}
 Experiments & Reward & Doors & MC & New \\ 
 \hline
 1 & $R_D$ & Single & Yes & No \\  
 2 & $R_D$ & Multi & Yes & No\\ 
 2 & $R_A$ & Single & Yes & No\\ 
 1 & $R_A$ & Multi & Yes & No\\ 
 1 & $R_A$ & Single & No & Yes\\ 
 1 & $R_A$ & Multi & No & Yes\\ 
 \label{tab:exps}
\end{tabular}
\end{center}



\noindent
These experiments can be grouped into two categories a) prior experiments being brought forth for new analysis and b) new experiments unique to this paper. Six of the PINSKY experiments fall into the former category while two fall into the latter. 

\subsubsection{New analysis of prior work}

Three experiments were run using reward function $R_D$. Two of these runs involved multiDoor dZelda, and the third focused on the standard singleDoor dZelda. 
Three experiments were run using the \emph{aligned} reward function $R_A$ (two singleDoor dZelda and one multiDoor dZelda). 

\subsubsection{New experiments}

New experiments using reward $R_A$ were run removing PINSKY's MC (one singleDoor and multiDoor) where it is hypothesized that the solve rate will decrease because the diversity of level types will collapse, thereby neutralizing inter-species transfer. 
Furthermore, the entire analysis methodology (explained next) that is being applied to each of the experiments is new to this paper. 



\subsection{Analysis Methodology}


The main contribution of this paper is an analysis of the transfer dynamics in PINSKY. Several types of analyses are proposed to examine PINSKY runs described in Section \ref{sec:dataset}. 
This work starts from the supposition that transfer is important \cite{wang:gecco19}\cite{dharna2020cogeneration}\cite{wang2020enhanced}. While the optimization loop is where agents learn to solve particular levels, the transfer step is where it is assumed that agents move between environments like a curriculum that emerges to increase overall agent performance. While transfer is permitted every $k$ (default=10) iterations of PINSKY, it is not guaranteed. As in \cite{wang:gecco19}, the first question this analysis asks is how many transfers occur during a run of the PINSKY algorithms.

Three analyses are performed: Analysis Total Transfer, Analysis To-and-From, and Analysis Blocks. Analysis \textbf{Total Transfer} looks at the importance of transfer through the lens of how often transfer occurs. 
The implication in POET is that the quality of transferred knowledge is affected by the similarity of the source and target tasks during the evolutionary process. Rather than being speciated during evolution, levels are clustered post-hoc through a method called cosine speciation (explained below), which is inspired by novelty search \cite{LehmanLocalComp} and described in Algorithm \ref{alg:cosineSpeciation}. Analysis \textbf{To-and-From} reexamines total transfer honing in on what types of transfer are happening and when (i.e. transfer at time $t$ from species $x$ to species $y$). Does transfer happen mostly within environment clusters? While identical transfer curves into and out of a cluster would suggest that most transfer occurs between the most similar levels, Analysis \textbf{Blocks} transitions from an agent-centric scale to a population-wide atemporal scale looking at the amount of transfer into and out of each class throughout PINSKY's evolutionary process by calculating probabilities of transferring between each class. By combining these transfer probabilities with additional information about whether and when the levels were solved, through Bayes Rule this analysis finishes by addressing the likelihood a level was solved given exposure to inter-species transfer (explained below). In this context, an agent transferring into a different species than the species it is currently optimizing for is the agent visiting a novel stepping stone. In POET this claim is assumed. 


\begin{algorithm}[h!]
  \DontPrintSemicolon
  \SetAlgoLined
  Create an empty set, A\\
  Pick a similarity threshold, $\gamma$\\
  Pick a set of vectors \textbf{V} $\in \mathbb{R}^{N}$\\
  Normalize the vectors to unit length\\
  \For{each vector, $\emph{v} \in \textbf{V}$}{
    \tcp{$\langle\cdot, \cdot\rangle$ is the dot-product in $\mathbb{R}^{N}$}
    \If{$\langle v, a \rangle < \gamma \text{, } \forall \text{ }  a \in A$}{
      \tcp{if A is empty, the comparison on line 6 is necessarily true}
      A.insert(v)\\
      \tcp{Assign v a color/ID}
    }
  }
  Return A \tcp{The set of unique species representatives}
  \caption{Cosine Speciation}
  \label{alg:cosineSpeciation}
\end{algorithm} 

\begin{algorithm}[h!]
\SetAlgoLined
\DontPrintSemicolon
  \SetKwInOut{Input}{input}
  
  \Input{$\mathbf{L}$: PINSKY levels. \newline 
  $\mathbf{\tau}$: Agent Transfer data \{(from\_id, to\_id, t), ...\}. \newline
  $\gamma$ (0.85): Cosine Similarity threshold}
  
  \tcp{Calculate level embedding}
  \For{ $ l \in L \text{ map to } m \in M$}{
    $x_1, x_2, x_3, x_4$ = Count unique game objects\\
    $x_5$ = Calculate $A^{*}$ path length from agent to key\\ 
    $x_6$ = Calculate $A^{*}$ path length from key to door(s)\\
    m = $\langle x_1,x_2,x_3,x_4,x_5,x_6\rangle$ \tcp{Fig \ref{fig:lvlCharacterization}}
  }
    
    Determine species representatives, A, via cosine speciation of $\mathbf{M}$ \tcp{Alg. \ref{alg:cosineSpeciation} and Fig. \ref{fig:manyDoorsProto}}
    
    \tcp{Classify each level into a species; Fig \ref{fig:578Sample}}
    \For{$m \in \text{normalized}(\mathbf{M})$ map to $s \in \mathbf{S}$}{
      s = argmax($A^{T} m$) 
      \tcp{This maps m.id to the nearest species id}
    }
    
    Calculate support for each species \tcp{Fig \ref{fig:lvlBreakdown}}
    
    Sort $\tau$ in temporal order\\
    Count successful transfers for each time t \tcp{Fig \ref{fig:globalTransferAllExps}}
    Count successful transfers into and from each species for each time t \tcp{Fig \ref{fig:manyDoorsTransferBreakdown}}
    Initialize empty transfer matrix, T\\
    \For{token $\in \tau$}{
      from\_id, to\_id, t = token\\
      T[S(from\_id)][S(to\_id)] += 1\\
    }
    T = T / total\_transfers \tcp{Fig \ref{fig:matrices}}
    \tcp{Everything not on T's diagonal is inter-species transfer.}
    
    Calculate the probability that a level was solved given inter-species transfer via Bayes Theorem: P(IST $|$ solved) / P(solved) \tcp{Table \ref{table:interSpeciesTransferGivenSolved}}

    Visualize agent transfers between the meta-population \tcp{Fig \ref{fig:manyDoorsTree}}
    
  \caption{Analysis of levels and agent transfer data}
  \label{alg:pinskyAnalysis}
\end{algorithm}

\subsection{Speciation Methodology}

Often as a way to protect innovation and prevent premature convergence, evolutionary algorithms niche or speciate the population such that individuals compete for survival in their own \emph{species} 
\cite{Stanley_evolvingneural}. While typical methods diversify based on genotypic representations, \emph{behavioral} speciation is another promising method to preserve diversity \cite{lehman2011abandoning}\cite{mouret2015illuminating}\cite{LehmanLocalComp}. While speciation occurs during evolution to protect promising individuals who have yet to meet fitness requirements of the general population, the approach of this paper is analogous to unsupervised clustering, where levels are separated into distinct groups after evolution completes, not during PINSKY's generation process. 

While novelty search creates a new species when a candidate individual exceeds a distance threshold from individuals in the archive, new species are created for PINSKY data when a candidate individual exceeds a uniqueness threshold based on similarity. While in novelty search the individuals must exceed a sparseness threshold of maximum distance from nearby individuals, the uniqueness threshold $\gamma$ for cosine speciation is the minimal cosine similarity between a candidate individual and the other already discovered clusters. 

\begin{figure}
    \centering
    \includegraphics[width=0.25\textwidth]{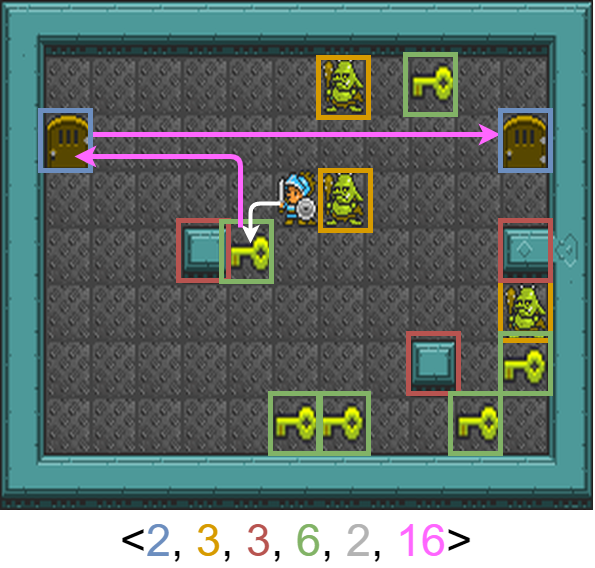}
    \caption{\textbf{Example level characterization vector.} 
    Semantic level information is compressed into six dimensions. In order of appearance in the vector: 2 doors, 3 monsters, 3 wall fragments, 6 keys, 2 steps from agent starting position to nearest key, and 16 steps from the nearest key to then visit each door. This phenotype-based level characterization creates an expressive embedding in $\mathbb{N}^6$.}
    \label{fig:lvlCharacterization}
\end{figure}

To calculate uniqueness of levels, they are first characterized as vectors by extracting their salient features as shown in Figure \ref{fig:lvlCharacterization}. The components extracted are features of the levels that are core to game play and subsequently used to obtain a similarity score. The first four components of a level vector are the numbers of doors, monsters, internal wall tiles, and keys it contains. The fifth and sixth components contain spatial information about a level's layout based on the A* path length between objects using the $L_1$ (Manhattan) distance admissible heuristic. The fifth and sixth components contain spatial information about a level's layout. The fifth component is the cost of the path an A* agent calculates from its starting position to the nearest key. The sixth component is determined by calculating the A* distance from the key in the fifth component to the nearest door such that $x_6 = $ A*(key1,door1). If there are multiple doors, then $x_6$ = A*(key1,door1) + A*(door1,door2) + ... where door2 is the closest from door1. 
Because the fifth and sixth components are based on the behavior of the A* agent, they not only consider distance between tiles but also indirectly measure the difficulty an agent faces when navigating mazes and provide a best-case baseline/upper-bound of agent efficiency regarding how to solve the problem.

While there are many approaches to determining similarity between vectors, this work explores using cosine similarity, the cosine of the angle between two (unit) vectors \cite{Sidorov2013}\cite{romesburg2004cluster}, to differentiate and cluster levels.
That is, given two level vectors $v_1, v_2 \in \mathbb{R}^N$, similarity, $\gamma$, is calculated as the projection of $v_1$ onto $v_2$; if $v_1,v_2$ are unit vectors (i.e. normalized by projecting them onto the Gauss sphere), then the \emph{cosine similarity} $\gamma \in [-1, 1]$.\footnote{A cosine similarity value of -1 denotes that the vectors are parallel but pointing in opposing directions; 0 denotes that the vectors are orthogonal; +1 denotes the vectors are parallel and pointing in the same direction.}  However, because the levels in this analysis are embedded as vectors in $\mathbb{N}^6$, cosine similarity for these vectors ranges between $(0,1]$ rather than $[-1,1]$, where higher values indicate more similarity.

In the data, a moderate uniqueness threshold for similarity is set as $\gamma$ = 0.85. This parameter was experimentally explored within $[0.5, 0.99]$ finding viable settings between $[0.6, 0.9]$. Interestingly when $\gamma \leq 0.6$, there were about two clusters (conflating observably distinct levels as the same species), and when $\gamma \geq 0.90$ the number of species doubled (and levels in each cluster were not distinct). Future work will investigate the impact of these settings on the analysis of transfer.

It should be noted that novelty-search-inspired methodologies of finding and determining species have a temporal dependence; the order in which level vectors are examined will change what levels get selected as representatives of their niches/classes. Whenever a new species is discovered or once every vector has been examined, the existing levels can be reclassified into their most similar species. For example, the 120th level might be best described by the species whose exemplar is the 824th level since level 824 was the first one to escape the cosine similarity boundaries of previously discovered levels in the archive.

While experiments were only performed for the game dZelda and analyzed by injecting some domain knowledge into the vectorized characterization of level, because games in GVGAI are all defined in VGDL, similar projections can be automatically determined for any of its games \cite{schaul2014extensible}.

\begin{figure*}
  \centering
  \subfloat[Species Representative levels (i.e. the first level to escape the cosine similarity regions of all already existing species representatives) for the first multiDoor dZelda experiment. The color above each level corresponds to distinct species.]{\includegraphics[width=.95\textwidth]{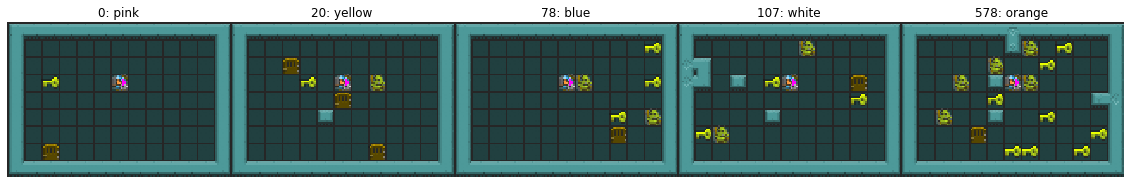}
  \label{fig:manyDoorsProto}}
  \hfill
  \subfloat[Levels were randomly sampled from multiDoor dZelda experiment 1 species 578 to examine if cosine-similarity resulted in similar levels being in the same species. Cosine-similarity for each level  is compared against level 578.]{\includegraphics[width=0.95\textwidth]{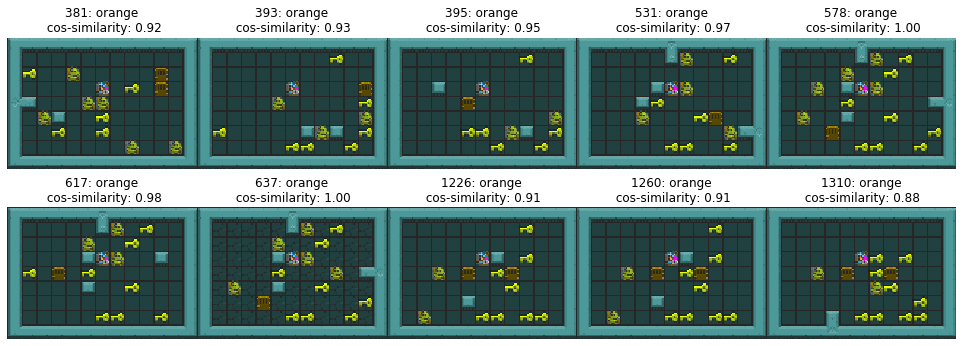}
    \label{fig:578Sample}}
    \caption{\textbf{Prototypical levels (i.e. species representatives) for multiDoor dZelda experiment 1 and a random sampling of levels belonging to species 578} showing that the cosine similarity speciation is finding clusters which are different across species (\ref{fig:manyDoorsProto}) and similar within a species (\ref{fig:578Sample}).}
    \label{fig:ClusteringVisual}
\end{figure*}





While this level characterization is expressive and can create qualitatively different clusters of levels while maintaining intra-cluster similarity (Figure \ref{fig:ClusteringVisual}), it is worthwhile to note that it is still only an approximation of the level's complexity. In Figure \ref{fig:578Sample}, levels 578 and 637 have a cosine similarity score of 1, indicating that the two levels are functionally the same. Upon inspection, this claim seems valid because the only difference between the maps is the placement of one of the chaser enemies from the upper right side of the map in level 578 to the lower left side of the map in level 637. Despite the overall similarity, changing monster placement can make two levels with similar characterization vastly different in difficulty (e.g. levels where the agent starts directly next to a monster versus the monster starting far away). 

Through a post-hoc analysis of cosine similarity, all of the levels generated by PINSKY are clustered and labeled by an ID determined by Algorithm \ref{alg:cosineSpeciation}. During the transfer period described in Algorithm \ref{alg:transfer}, an agent, $a_i$ in the active meta-population of pairs is free to transfer to different level, $e_j$ if $a_i$ scores better on $e_j$ than all other agents in the active meta-population. A core premise in both POET and PINSKY is that transfer is an important mechanism for increasing agent performance and task generation. 

\section{Results}
\begin{figure}
    \centering
    \includegraphics[width=0.45\textwidth]{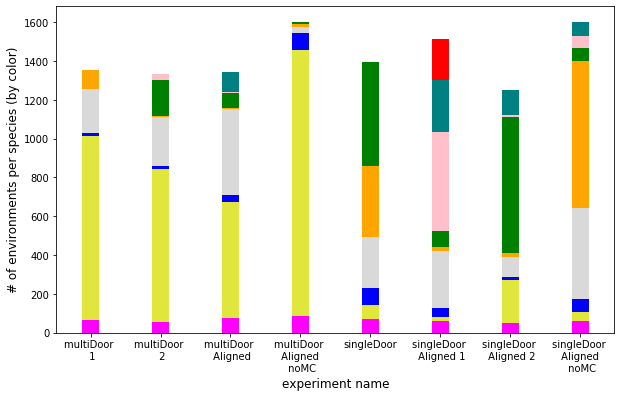}
    \caption{\textbf{Level distribution by species.} Often a single species of levels dominate over other species, but PINSKY manages to generate levels that require diverse behaviors to solve. Species are colored in order of emergence the initial species being pink, then yellow, blue, white, orange, etc.}
    \label{fig:lvlBreakdown}
\end{figure}

The goal of this analysis is to gain insight about the role of inter-species transfer in POET-inspired coevolutionary systems. By applying the cosine similarity speciation and taking into account the temporal nature of the level generation 
(i.e. comparing the environments by order of creation), the final distribution of level classification can be observed in Figure \ref{fig:lvlBreakdown}. For many of the experiments (particularly with multiDoor dZelda), often a single species of levels dominate over other species, and this trend is exasperated when the MC is removed. 

\begin{figure}
    \centering
    \includegraphics[width=0.48\textwidth]{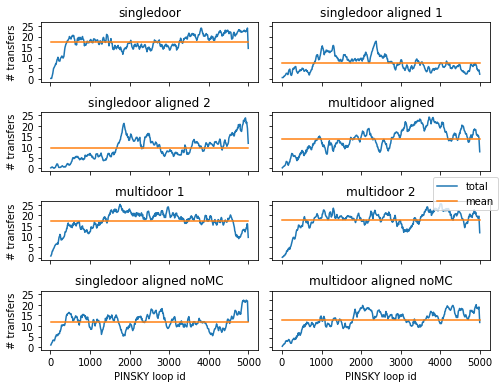}
    \caption{\textbf{Analysis Total Transfer / Smoothed total transfer curves for each experiment}. Transfer dynamics of aligned vs non-aligned environments show different long-term transfer patterns where on average in the aligned environments, transfer is less prevalent, p $< <$ 0.05.}
    \label{fig:globalTransferAllExps}
\end{figure}

\begin{figure}[ht!]
    \centering
    \subfloat[Breakdown of transfer dynamics \textbf{to species X} plotted temporally.]{\includegraphics[width=0.48\textwidth]{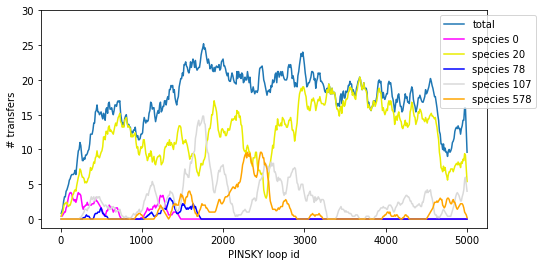}\label{fig:manyDoorsTransferTo}}
    \hfill
    \subfloat[Breakdown of transfer dynamics \textbf{from species X} plotted temporally.]{\includegraphics[width=0.48\textwidth]{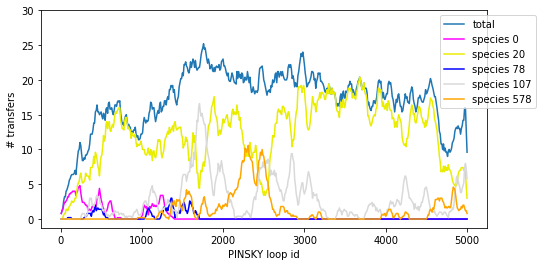}\label{fig:manyDoorsTransferFrom}}
    \hfill
    \subfloat[Negative values suggest transfer out of the species; positive values into the species.]{\includegraphics[width=0.48\textwidth]{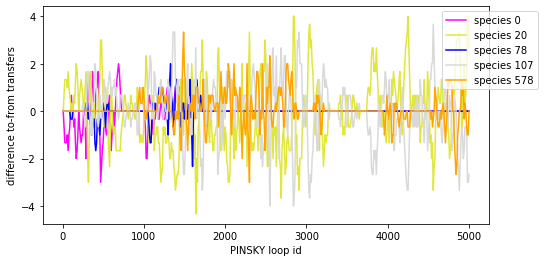}\label{fig:manydoorsTransferDifference}}
    \hfill
    \caption{\textbf{Analysis To-and-From / Total lifetime transfer dynamics for the multiDoor dZelda 1 experiment by species.} Interestingly, non-dominant species wax and wane in relevance over time. Furthermore, the transfer-to and transfer-from curves are not one-to-one, suggesting that agents are in fact transferring between species.}
    \label{fig:manyDoorsTransferBreakdown}
\end{figure}

\begin{figure}
    \centering
    \subfloat[singleDoor]{\includegraphics[width=0.24\textwidth]{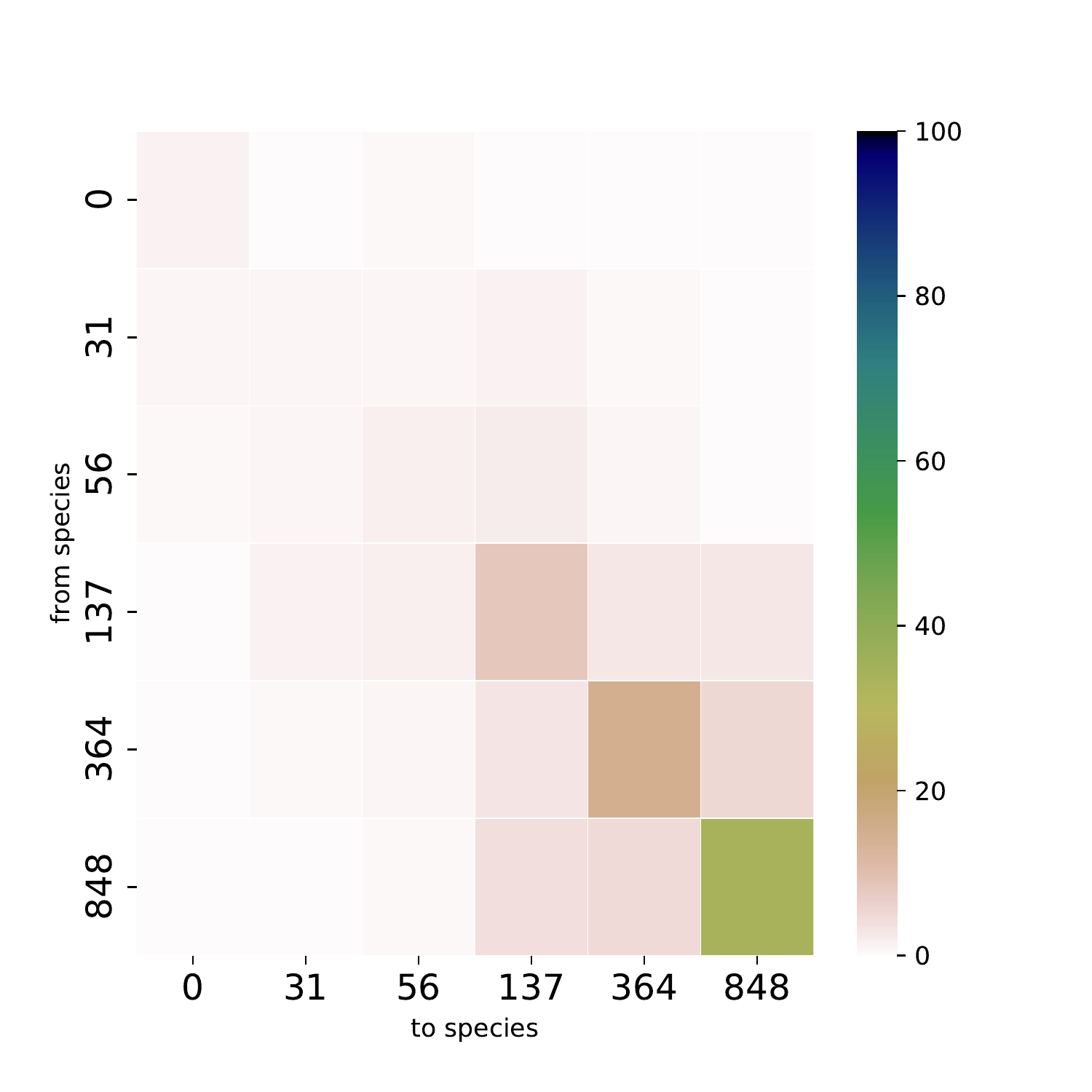}\label{fig:singleDoorMatrixTransfer}}
    \hfill
    \subfloat[multiDoor 1]{\includegraphics[width=0.24\textwidth]{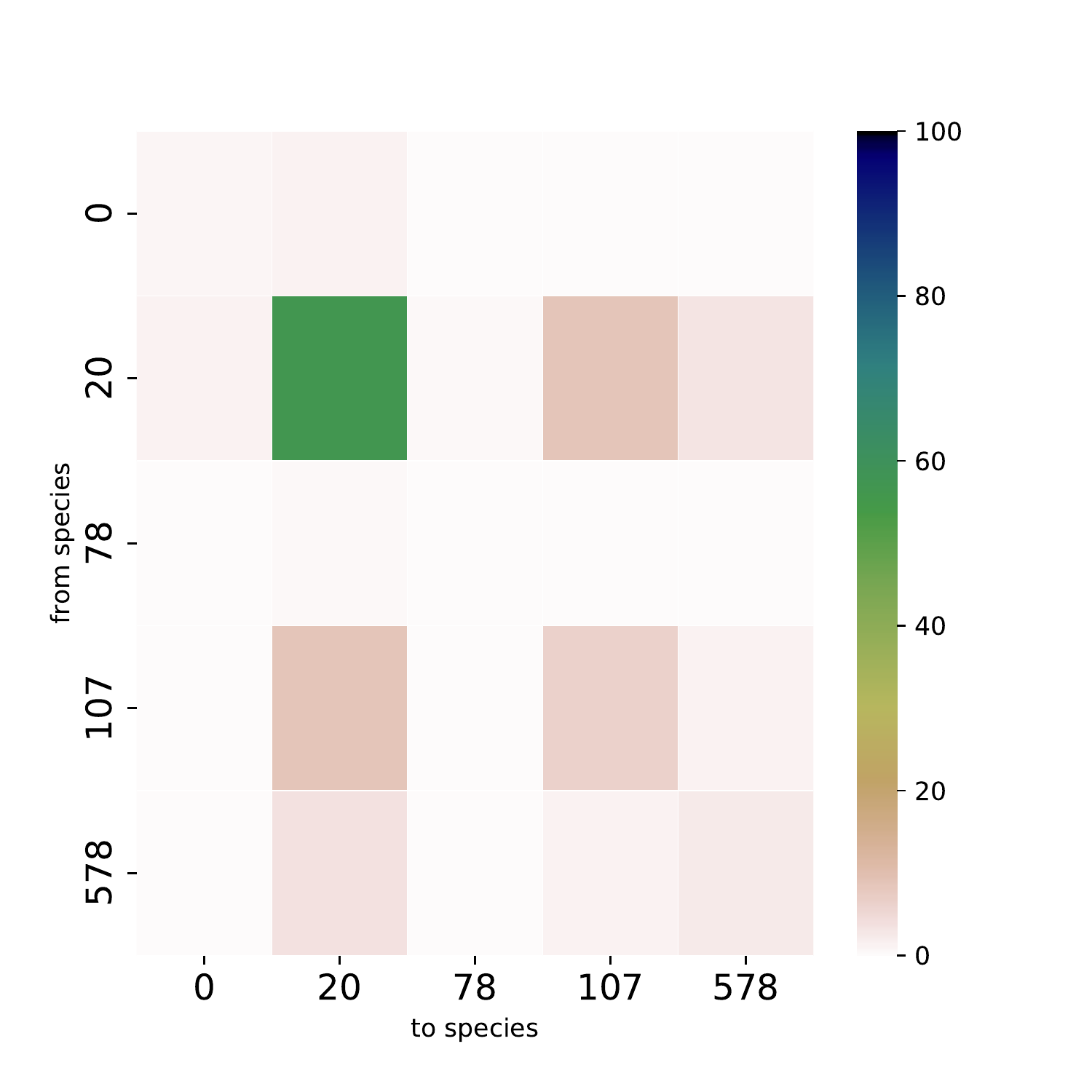}\label{fig:manyDoorsMatrixTransfer}}
    \hfill
    \subfloat[multiDoor 2]{\includegraphics[width=0.24\textwidth]{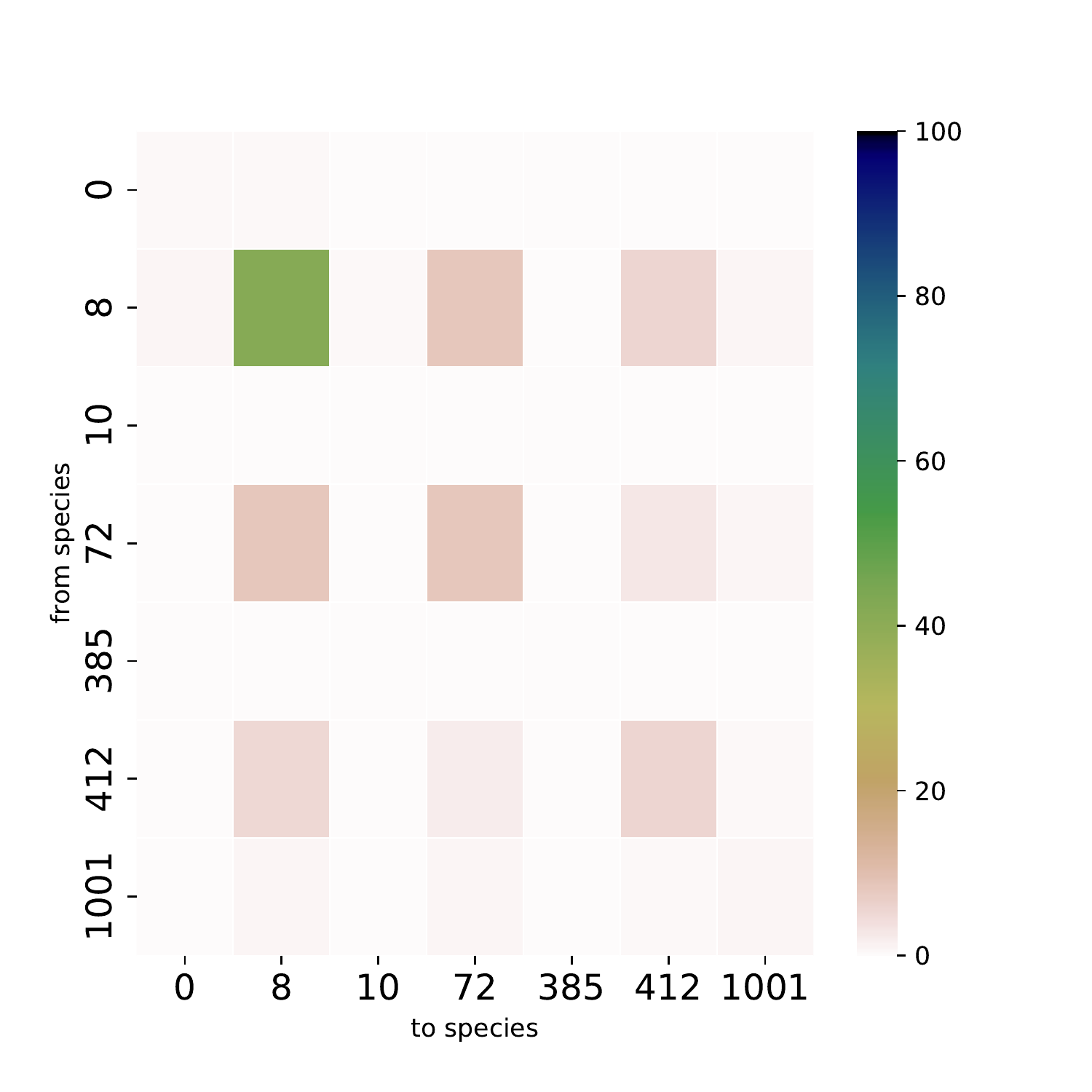}\label{fig:manyDoors2MatrixTransfer}}
    \subfloat[singleDoor Aligned 1 ]{\includegraphics[width=0.24\textwidth]{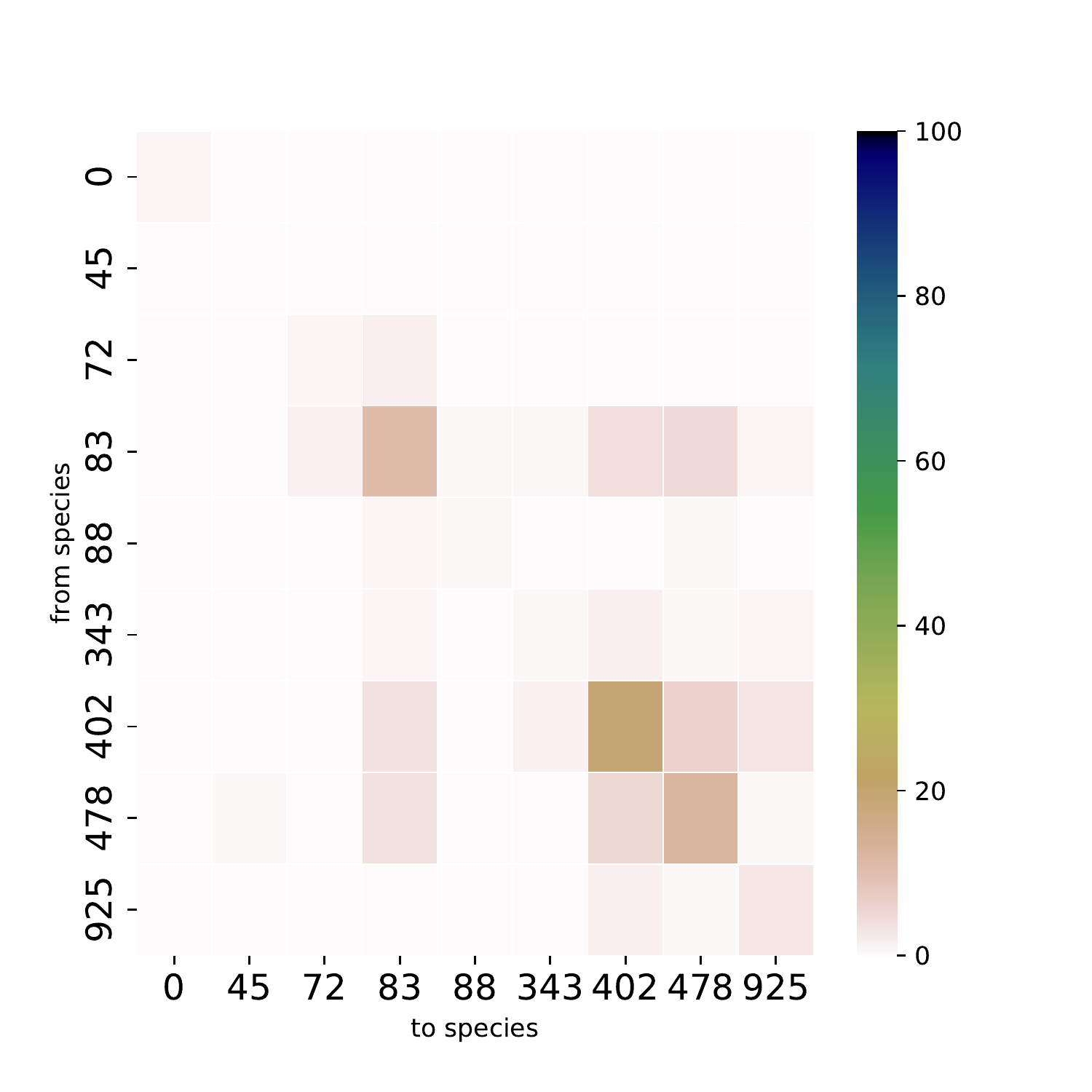}\label{fig:singleDoorAligned1MatrixTransfer}}
    \hfill
    \subfloat[singleDoor Aligned 2]{\includegraphics[width=0.24\textwidth]{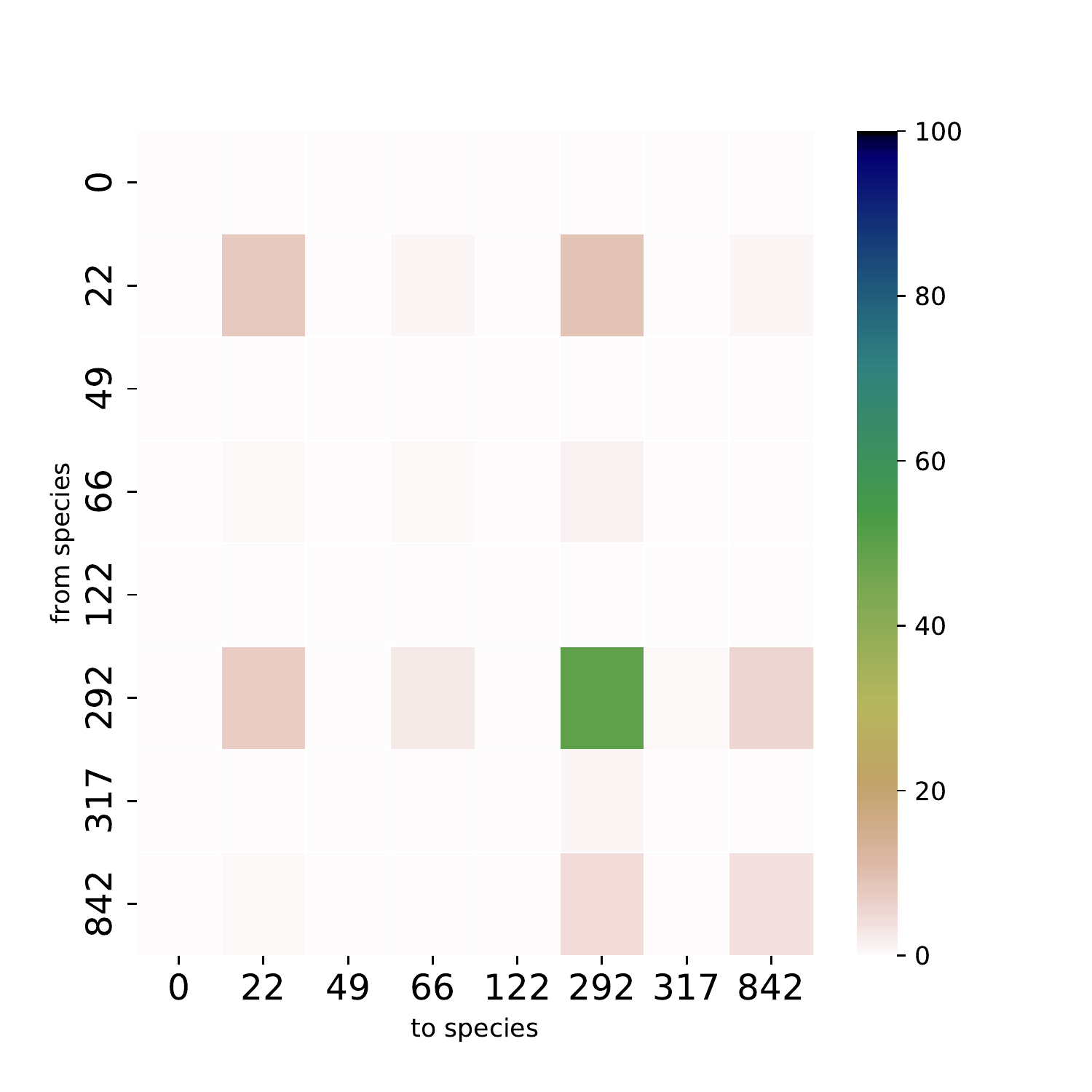}\label{fig:singleDoorAligned2MatrixTransfer}}
    \hfill
    \subfloat[multiDoor Aligned]{\includegraphics[width=0.24\textwidth]{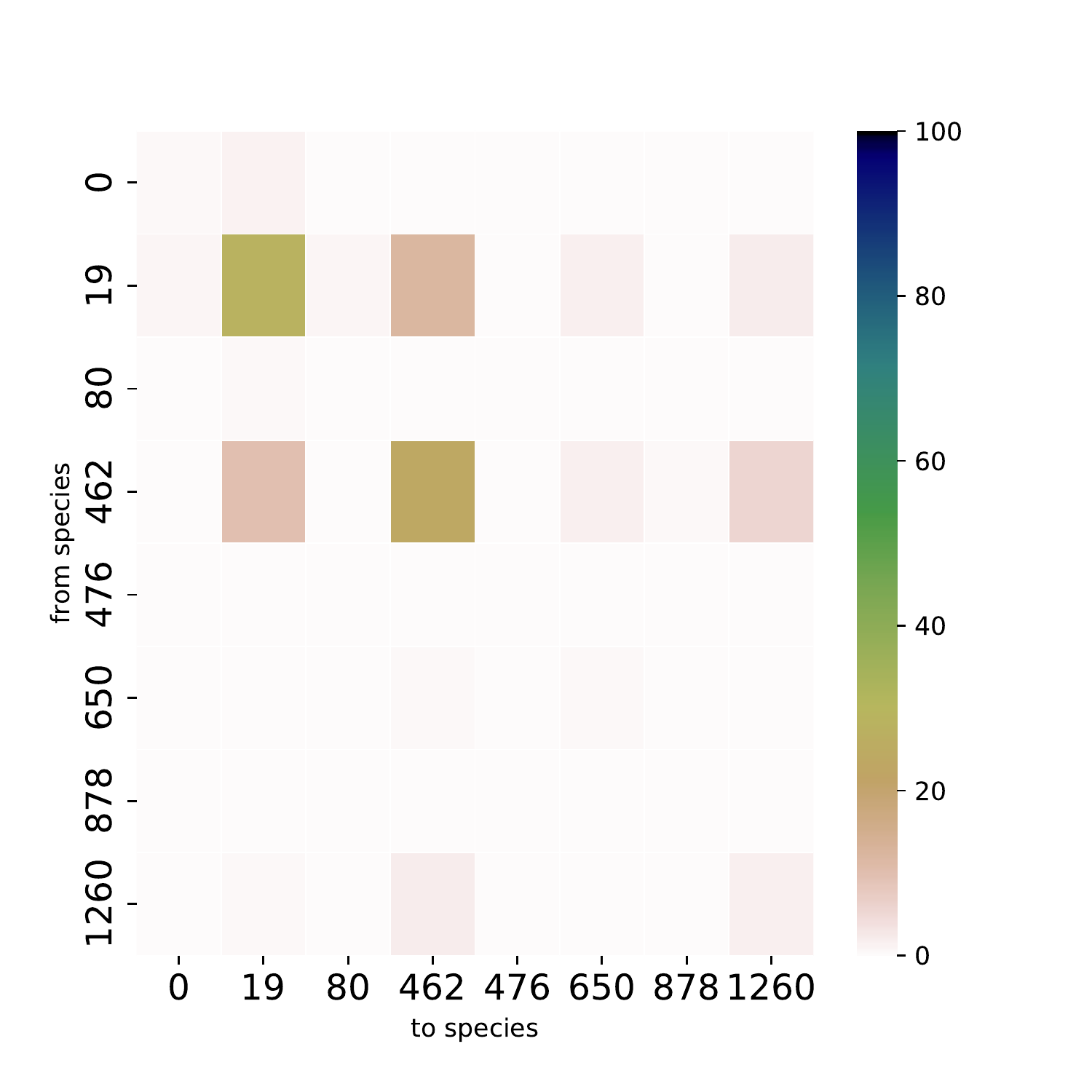}\label{fig:manyDoorsAlignedMatrixTransfer}}
    \hfill
    \subfloat[singleDoor aligned noMC]{\includegraphics[width=0.24\textwidth]{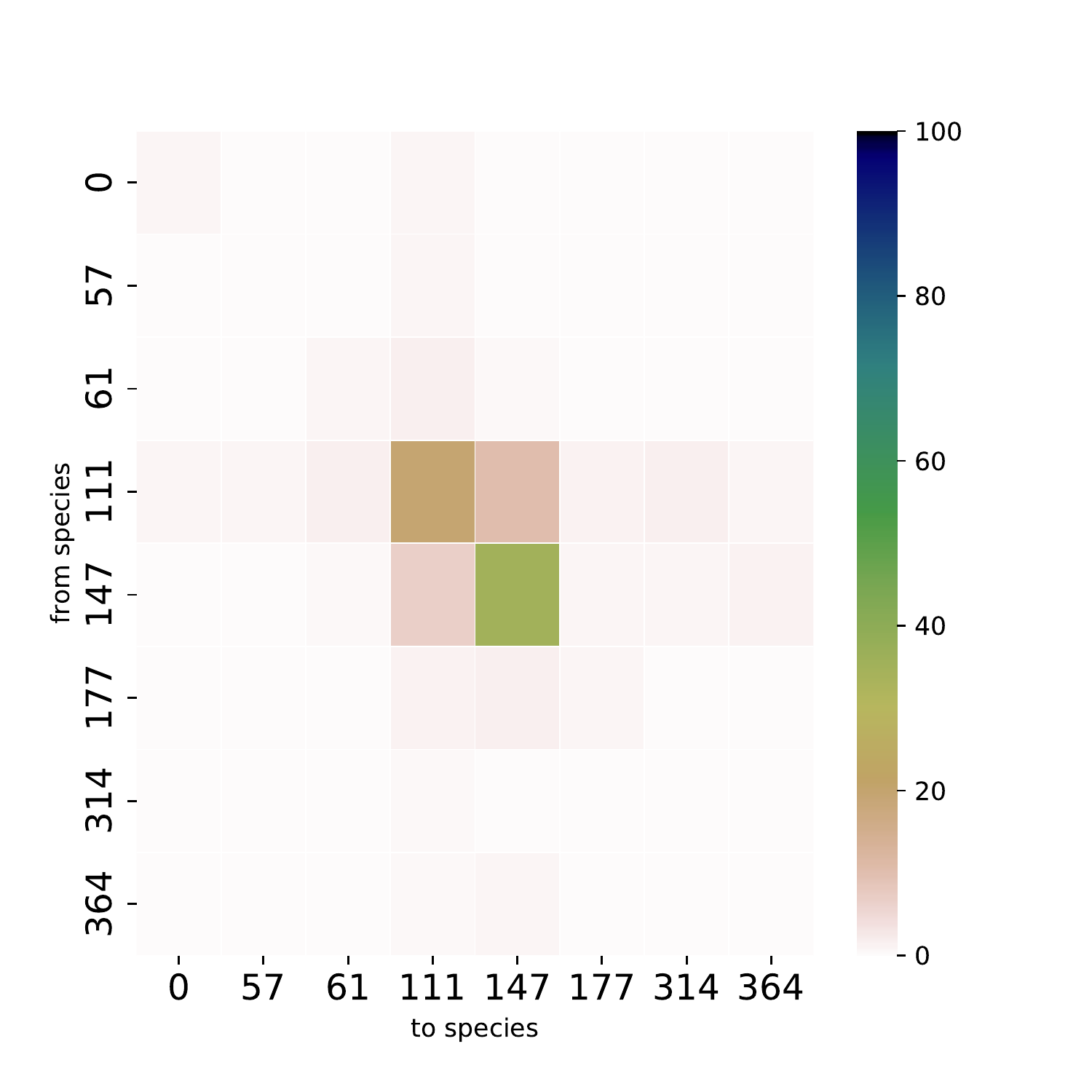}}\label{fig:noMCSingleAligned}
    \hfill
    \subfloat[multiDoor aligned noMC]{\includegraphics[width=0.24\textwidth]{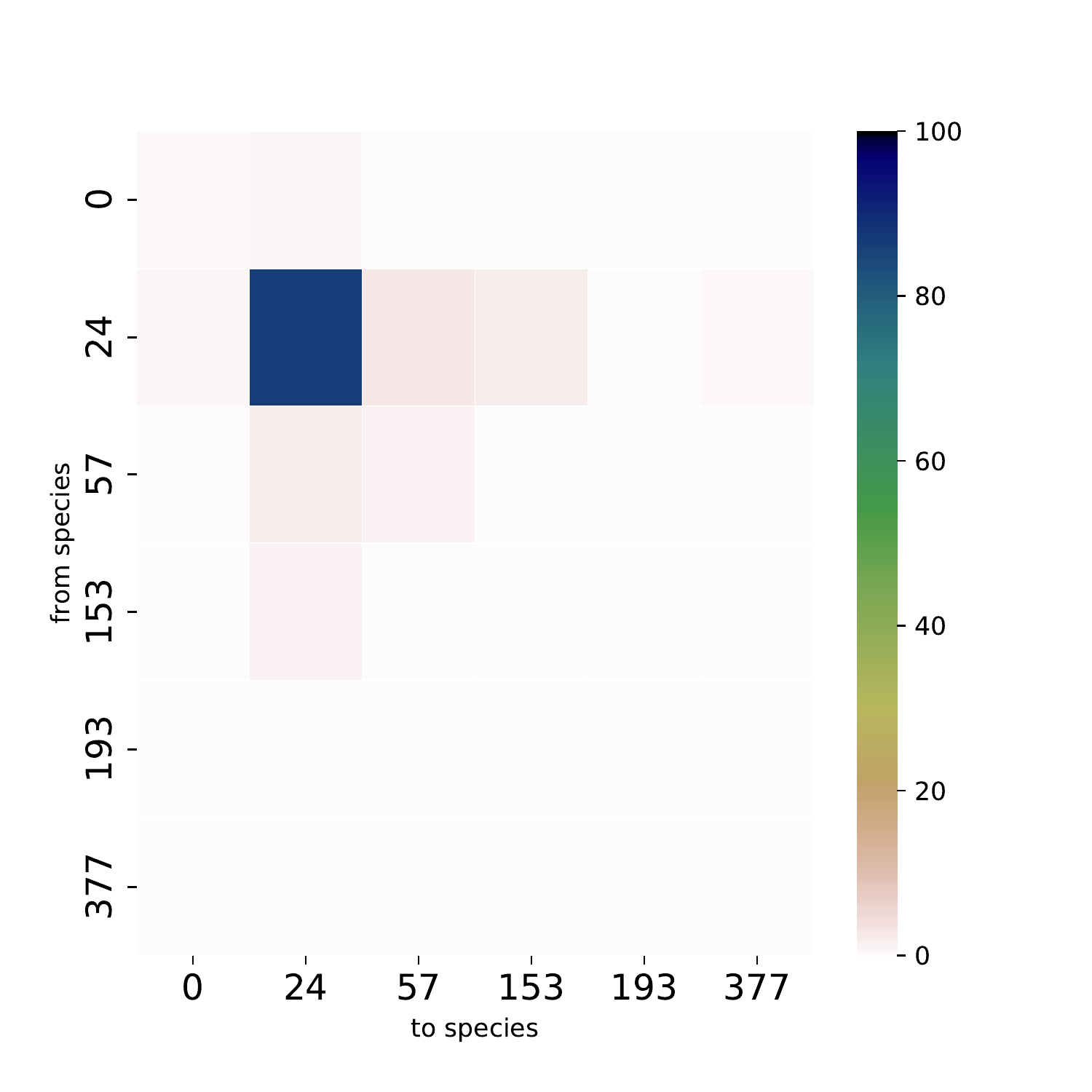}}\label{fig:noMCMultiAligned}
    \caption{\textbf{Analysis Blocks / Inter- and intra-species transfer percentages for dZelda games.} Y-axis is the species the agent transferred from and the X-axis is the species the agent transferred to. Note that the off-diagonal represents inter-species transfer. If all transfer was intra-species transfer, all off-diagonal entries would be zero and the to- and from-transfer graphs in Fig \ref{fig:manyDoorsTransferBreakdown} would be identical.}
    \label{fig:matrices}
\end{figure}

Levels are classified according to Algorithm \ref{alg:cosineSpeciation}, and then support is calculated for each class (Figure \ref{fig:lvlBreakdown}). Raw transfer curves for each experiment can be seen in Figure \ref{fig:globalTransferAllExps}. The raw curves can also be broken down per class to observe the waxing and waning of each class over evolutionary time and visualize transfer between the classes as seen in Figure \ref{fig:manyDoorsTransferBreakdown}. Using the classified agent-env pairs, 
the transfer data generated by PINSKY can be used to determine the class of each agent that enters or leaves a given pair. These agent updates provide a measure of whether agent transfer is happening between pairs that share a classification or pairs that are of different classes i.e. number of transfers from other classes divided by number of transfer. Figure \ref{fig:matrices} transitions from an agent-centric scale to a population-wide atemporal scale looking at the amount of transfer into and out of each class as a whole used to calculate probabilities of transferring between each class. Algorithm \ref{alg:pinskyAnalysis} gives a full breakdown of the analysis in this paper. 

Figure \ref{fig:globalTransferAllExps} shows the raw number of transfers as they happen over the course of a PINSKY run as well as the mean number of transfers for each experiment. On average, the unaligned experiments exhibited greater amounts of agent transfer than the aligned experiments. Means tests (Mann-Whitney U-Tests and Relative T-Tests) found a significant difference ($p < < 0.05$) between the transfer dynamics in the experiments providing evidence that different reward functions and minimal criteria meaningfully change the transfer dynamics (Figure  \ref{fig:globalTransferAllExps}). However, Figure \ref{fig:globalTransferAllExps} does not take into account the speciation and instead shows global trends of how much PINSKY relies on the transfer mechanism. Therefore, the first natural question that arises from speciating the levels is: does transfer happen only within a species, or does it also occur between species?  

As shown in Figure \ref{fig:lvlBreakdown}, a single species often dominates the distribution of different level types, leading to the conjecture that one species should also account for a majority of all transfers. However, by looking at the transfer curves (Figure \ref{fig:globalTransferAllExps}) and breaking them down into their respective species, Figure \ref{fig:manyDoorsTransferBreakdown} shows that even late in the evolutionary run, a species can return to relevance after having been neglected for a period of time. Furthermore, throughout the evolutionary process, individuals are migrating between different species of tasks (Figures \ref{fig:manyDoorsTree} and \ref{fig:manydoorsTransferDifference}). For example, in the first multiDoor dZelda experiment, \emph{all} transfers briefly happen purely among species 20. However, individuals in the other species do not just die off. 
In the multiDoor 1 experiment, 66\% of all transfer happens within species, with 56\% of all transfers happening within species 20 (Figure \ref{fig:manyDoorsMatrixTransfer}). 50+\% of transfers happen between members of the same species and a large portion of the remaining 50\% happens between the first-and-second-most populous species. 

Previous experiments \cite{dharna2020cogeneration} demonstrated that the \emph{aligned} reward function dramatically improved the number of levels PINSKY created and solved (Table \ref{table:interSpeciesTransferGivenSolved}). 
The transfer curves in Figure \ref{fig:globalTransferAllExps} provide some additional clarity as to how PINSKY achieved this feat. For the first singleDoor experiment, its total transfer numbers were around 20 transfers each transfer step. However, the singleDoor aligned experiment transfer curves saw a decrease in transfers over time with the second singleDoor aligned experiment seeing a decrease in transfers followed by a subsequent increase. 

When the reward alignment is left in place, but 
PINSKY's MC is removed, the solve rate returns to its pre-aligned levels. For the noMC experiments, Figures \ref{fig:lvlBreakdown} and \ref{fig:matrices} show that a) most of PINSKY's generated levels are similar (i.e. level diversity has collapsed) and b) that the vast majority of the transfer is within these mega-species which means that inter-species transfer has been impeded as hypothesized. Combined, these observations suggest that the MC was helpful for fostering diversity in the generated levels which was then exploitable by PINSKY's transfer mechanism to help train high-performing agents. 

Figures \ref{fig:manyDoorsTransferBreakdown} and \ref{fig:matrices} suggest that transfer among a single species is the primary mode of transfer in PINSKY and therefore responsible for the system's ability to create and solve difficult problems. However, Table \ref{table:interSpeciesTransferGivenSolved} shows that a minimum of 39\% of solved levels required inter-species transfer, and a maximum amount of 75\%.\footnote{The noMC cases are not included in this comparison since removing the MC collapses level diversity and impedes inter-species transfer therefore the comparison is not fair.} Despite 83\% of all transfer happening inside species 24 in the multiDoor noMC experiment (Figure \ref{fig:matrices}), 31\% of solved levels exhibited inter-species transfer (Table \ref{table:interSpeciesTransferGivenSolved}). These conditional probabilities show the necessity of inter-species transfer more than Figures \ref{fig:manyDoorsTransferBreakdown} and \ref{fig:matrices} imply alone. 


It is interesting to note that if a generalist agent (i.e. one that solves all or most created levels) were to be trained, then that agent would transfer into each level and therefore the transfer value would be maximized at 30 (the meta-population size) and then drop to zero as the generalist agent would not be able to be defeated (by a copy of itself living in other environments). That does not happen. Instead, agents specialize; as POET is solving increasingly complex levels the transfer curves reduce, showing that it becomes harder for an incumbent agent to be replaced (Figures \ref{fig:globalTransferAllExps} and \ref{fig:manyDoorsTransferBreakdown}).

\begin{table}[htbp]
  \centering
  \begin{tabular}{ |p{2.6cm}|p{2cm}|p{3cm}| }
    \hline
    Experiment & \% of viable solved levels & p(solved$ | $inter-species transfer)\\
    \hline \hline
    singleDoor        & 62 & 0.74\\
    singleDoor aligned 1   & 90 & 0.39\\
    singleDoor aligned 2   & 83 & 0.41\\
    singleDoor alg. noMC & 61* & 0.41\\
    multiDoor 1       & 13 & 0.75\\
    multiDoor 2       & 8 & 0.49\\
    multiDoor aligned & 29 & 0.47\\
    multiDoor alg. noMC & 10* & 0.31\\

 \hline
    \end{tabular}
  \caption{\textbf{Percentage of solved viable levels for each experiment as well as the probability that a level was solved given transfer from an outside species.} Notably, the limited success of PINSKY on difficult domains (e.g. multiDoor dZelda) is not-insignificantly due to inter-species transfer. Column II is calculated using Bayes Rule. \newline *not all levels might be viable because the MC was removed.}
    \label{table:interSpeciesTransferGivenSolved}
\end{table}

Figure \ref{fig:manyDoorsTree} visualizes the phylogenetic tree 
of the first 105 environments of PINSKY's meta-population. The black edges between nodes indicate the phylogenetic ancestry or lineage of an agent-environment pair, starting from a randomly initialized agent and a seed level for the generator (shown in Figure \ref{fig:manyDoorsProto} level 0). Node colors indicate the species of a pair, where species is determined through Algorithm \ref{alg:cosineSpeciation}. Outgoing red edges like those from node 75 to 104 indicate that an agent of one pair has successfully transferred to another environment. Using the Hardcore Bipedal Walker domain, Wang et al. suggest that close relatives are more genetically related (i.e., similar) than those with only distant ancestors \cite{wang:gecco19}\cite{wang2020enhanced} and that transfer between nodes with distant ancestors is important for discovering high-performing individuals. Figure \ref{fig:manyDoorsTree} shows that similar environments do not necessarily descend from similar ancestor pairs. For example, blue species emerge from both pink and yellow parents (e.g., pairs 59 to 78, 4 to 32, and 69 to 93), while pink and yellow species can emerge from each other (e.g., 3 to 7, 7 to 29). This result is not surprising because of the level embedding’s limited expressivity and inability to capture geometric information.

Although, as shown earlier, most transfers happen between levels in the same species (indicated by color), there are clear examples of inter-species transfer between nodes in Figure \ref{fig:manyDoorsTree} with distant ancestors (e.g. level 102's agent (yellow, bottom center) transfers into level 81 or when level 93's agent (blue, center) transfers into level 103) as well as transfer between nearby but different species of nodes (level 103 to 97). The cosine similarity between levels 102 and 81 is 0.57, 93 and 103 is 0.71, and 103 and 97 is 0.73. 

An agent that has specialized into a particular environment is likely to fail when presented with a more difficult environment. However, agents transferring \emph{backward} (from a more difficult environment into an easier environment) is observed in PINSKY.
Backwards transfer can be observed between levels 75 and 80 and is shown in Figure \ref{fig:manyDoorsTree}. Levels 75 and 80 are nearly identical, which makes sense as level 80 is the direct descendent of level 75. The difference between the two levels is that a monster's starting position has been moved from the right of the agent's starting position and placed directly in the shortest path from the agent's initial position to the key and then the goal. As PINSKY attempts to (independently) solve level 80, what it has learned is backward compatible with level 75 as the agent in level 80 proves able to replace level 75's agent. Notably, level 75 is solved only after level 80's agent transfers into environment 75, but level 80 remains unsolved for its entire duration in the PINSKY active population.

Agents transferring into more difficult environments can also be seen in Figure \ref{fig:manyDoorsTree}. Level 86's agent transfers into level 92. These two levels have a cosine-similarity of 0.95 and share some macro features (each level has two doors and a similar distance from the agent to key) so a transfer here seems reasonable. However neither environment was solved.

\section{Discussion}

\subsection{Evolutionary Dynamics}

Phylogenetic trees corresponding to biological evolution are sprawling, deep, and contain nested branches \cite{wang2020enhanced}. While researchers in the field of artificial life seek to replicate this phenomenon in artificial evolutionary systems, biologically-inspired generative algorithms often do not produce phylogenetic trees like those in nature.
Figures \ref{fig:lvlBreakdown}, \ref{fig:manyDoorsTransferBreakdown}, and \ref{fig:manyDoorsTree} imply that PINSKY's phylogenetic trees are also unlike those found in nature. In fact, Figure \ref{fig:lvlBreakdown} shows that an often overwhelming number of environments belong to a single species; however, Figure \ref{fig:manyDoorsTransferBreakdown} shows that the smaller species are still important and can reemerge. Furthermore, Table \ref{table:interSpeciesTransferGivenSolved} shows that different species play an important role in solving PINSKY-generated environments where many of the solved levels exhibit between-species transfer dynamics thereby navigating unlikely stepping stones on their path to solving difficult levels.

The term \emph{exaptation} describes a shift in the utilization of a trait \emph{during} evolution \cite{gould_vrba_1982}. Phrased another way, an innovation that occurs in a particular evolutionary context later becomes useful for a completely different purpose. We claim that transfer in PINSKY is strongly related to this relatively understudied phenomenon. When a better-suited agent is found for a level,
the better agent replaces the incumbent agent. The behavior which previously served the purpose of solving environmental challenge A now solves a new challenge, B.  

The tournament update inside POET evaluates each active agent 
in every other active environment to find the best agent for each environment. This population-wide competition instigates zero-shot evaluations of the agents on each level. If agents are rewarded for side quests (e.g. killing enemies), as when using $R_D$, then it makes sense that that transference of agents is more common. An agent that learned to kill one monster would be replaced by an agent that kills one monster and also picks up a key, or by an agent that kills two monsters (i.e. orthogonal agent improvement with respect to the overall goal of winning the game). However, simply killing two monsters brings the agent no closer to the winning game state and in fact reinforces the skills needed for killing instead of winning, so the ``improvement'' brought by replacing the incumbent agent is actually a degradation of the agent's ability. Degraded ability after transfer  is a phenomenon known as negative transfer \cite{Taylor09Negative}\cite{wang2018characterizing}.

One potential reason for the transfers decreasing could be that as the environments diverge from each other (i.e. more species are spawned or the levels come from lineages with different specialties), there is less immediate overlap between the skills needed to solve one level versus another (e.g. killing monsters because they block a critical path versus running away from monsters so as to not get caught). In the earlier stages of the PINSKY process, agents with specific behaviors might move between environments more easily. 

The \emph{aligned} reward, $R_A$, function makes the agent optimization into a process that can be more cleanly solved (i.e. the algorithms are not being tricked into moving into parts of the parameter space that give high reward but do not move the agent closer to winning the game). Furthermore, using the aligned reward function helps PINSKY avoid issues of negative transfer \cite{wang2018characterizing} caused by the distracting default GVGAI reward function. Therefore, in the cases where the reward/fitness function directly incentivizes the desired behavior, transfer becomes more meaningful and therefore less common. However, the cost of the aligned reward function is that it is sparse providing feedback only at the end of a rollout. 

In the singleDoor dZelda aligned 2 experiment, the transfer rate spiked, dipped, and then started to spike again (Figure \ref{fig:globalTransferAllExps}). The initial spike matches the behavior shown in the other aligned transfer graphs, but the secondary rise in transfers is a new phenomenon. While the agent networks might have specialized (thereby decreasing the amount of possible transfer), the majority of unique species contained levels not yet solved by the current crop of PINSKY agents. Therefore, transfer likely increases again to compensate for the harder tasks. 


Given that killing enemies is an important part of dZelda games, it makes sense that a designer might consider explicitly rewarding that behavior. However, the reward shaping inherent (e.g. killing monsters) in the default GVGAI reward function can overpower the sparse task reward (e.g. winning the game). As a result of learning with the reward shaping, the agent is not learning the desired task \cite{huang2020action}\cite{NgReward1999}. 

Once an agent in a reward-\emph{aligned} dZelda experiment has solved a level, the only way for it to be replaced by another agent, is if the prospective new agent can solve the level \emph{faster} than the incumbent agent. On the flip side, if the incumbent agent is killed by enemies, then it can be easily replaced by agents that survive longer before dying, agents that purely stay alive, or agents that actually solve the task. By definition of the \emph{aligned} reward function, solutions improve with each transfer until the minimum path solution is found. Since the reward function is truly indicative of whether or not the task has been accomplished, transfer only happens when meaningful improvement has been made by the new agent with respect to the incumbent agent. By \emph{removing} the ``distracting'' portions of the reward function, the optimization algorithm cannot move towards deceptive local optima as happens when using the standard dZelda reward function. Therefore, given that PINSKY's optimization steps are more limited in their direction of improvement, the transfer aspects of POET-style algorithms allow the population to search for behaviors outside of being driven purely by the optimization algorithm. In other words, PINSKY can find meaningful stepping stones throughout its population and transfer agents into the environment they thrive in, thereby giving the optimization algorithm a better foothold in the reward landscape from which to continue optimizing.

\subsection{Connecting Coevolution with Machine Learning}

The long-term dynamics of transfer learning in POET-like systems provides hints about similar coevolutionary phenomena. 
Critically, POET-style algorithms are not training under a minimax setup due the minimal criteria (MC). The MC loosens the constraint that the generative evolutionary process create environments that  minimize agent ability, and instead new environments that are simply ``good enough'' are allowed to exist. By not explicitly optimizing (i.e. minimizing) but instead providing a floor, the MC allows the generative process to create diverse stepping stone environments that the agents can use to solve difficult potentially orthogonal tasks. When the MC is removed from PINSKY, the level diversity collapses (Figure \ref{fig:lvlBreakdown}) mirroring mode-collapse in a GAN \cite{guttenberg2018potential}, and even with the reward being aligned, the amount of solved tasks drops. 
Needing both the reward alignment as well as an MC provides a baseline that future not-quite minimax (and in the case of no MC, minimax) training schemes can adopt. 


The presence of reward-alignment and transfer is mirrored in self-play RL settings such as AlphaStar \cite{vinyals2017starcraft}. AlphaStar maintains a population of agents to get diverse strategies, and when the agents play against each other, they participate in a zero-sum (i.e. well-aligned) game to determine which agent will live (transferring its weights to the loser). Whereas AlphaStar seeks to create a diverse initial set of agents using human data and maintain the diversity using highly-shaped reward functions specific to each cluster, integrating an MC might allow setups like AlphaStar to become entirely self-generative and remove the seeded human data without impacting the algorithm's overall effectiveness. 


\subsection{Connection to meta-learning}

Researchers have proposed meta-learning procedures that optimize a set of meta-parameters in an inner-outer loop training scheme \cite{alex2019evolvability}\cite{grefenstette2019generalized}\cite{finn2017modelagnostic}. When this paradigm is gradient-based, the nested structure allows inner-loop training to specialize a model for a task, calculate derivatives, and then backpropagate through the inner-loop learning algorithm to update the meta-parameters in the outer-loop  \cite{grefenstette2019generalized}\cite{finn2017modelagnostic}. Alternatively, POET-like systems contain a meta-population where each individual undergoes inner-loop optimization, and the population coevolves in response to the current structure and ability of agents in the meta-population. Upon finishing the inner-loop optimization, evaluation of each combination of each agent and environment in the meta-population is executed to update agent-environment pairings. In this sense, POET's meta-population fills a similar roll as meta-parameter models in the gradient-based optimization schemes mentioned earlier.

\subsection{Impact of limited generalization}

While PINSKY is able to create agents able to solve levels unsolved by agents evolved from scratch, it has not so far been able to find a single agent able to solve all levels. This fact, combined with other observations, suggests that the neural network architecture used might preclude truly general policies. This is in accord with recent work on the failure of reinforcement learning to generalize~\cite{justesen2018illuminating}. Agent-centric inputs to the agent might enable more general solutions~\cite{ye2020rotation}.

\subsection{Future Work}
 How can PINSKY engage in more meaningful transfer learning dynamics beyond an aligned, but sparse, reward function \cite{dharna2020cogeneration}? One potential answer is to explicitly embrace the meta-learning structure of POET. For example, soft RL optimization methods that explicitly maximize an agent's long-term reward as well as making the policy as general as possible \cite{haarnoja2018soft} can be explored as alternative inner-loop optimization processes. Similarly, one might use methods specifically designed to learn policies that are easily adaptable \cite{finn2017modelagnostic}\cite{song2020rapidly}\cite{alex2019evolvability}.
 Importantly, however, analysis suggests the need for more expressive environments \cite{yu2019metaworld} or environmental encodings \cite{Gaier2020Discovering} or to explore the space of current tasks more exhaustively focusing on finding greater diversity between levels \cite{mouret2015illuminating}. 



\section{Conclusion}

This paper shows that transfer (i.e. meta-population realignment) mostly happens between similar environments, but that inter-species transfer, although rarer, was integral to PINSKY system performance. This paper offers a novel analytical framework (i.e.\ post hoc speciation, phylogenetic tree visualization, species-sensitive transfer matrices) for studying long-term coevolutionary dynamics in POET-like systems through the lens of transfer learning. The analysis also casts PINSKY in the framework of generalized inner loop meta-learning algorithms \cite{grefenstette2019generalized}. 
Under these frameworks, we show how PINSKY manages to solve difficult environments through repeated policy transfers between different species of ``stepping stone'' environments ultimately resulting in specialized agents that can solve otherwise unsolvable PINSKY-generated tasks. Furthermore, we show the importance of minimal criteria on POET-like systems and its impact on transfer learning.







\section*{Acknowledgment}
This work was supported by the National Science Foundation. (Award number 1717324 - “RI: Small: General Intelligence through Algorithm Invention and Selection.”).

\ifCLASSOPTIONcaptionsoff
  \newpage
\fi



%

\bibliographystyle{IEEEtran}
\bibliography{bibliography} 

\begin{thebibliography}{10}
\providecommand{\url}[1]{#1}
\csname url@samestyle\endcsname
\providecommand{\newblock}{\relax}
\providecommand{\bibinfo}[2]{#2}
\providecommand{\BIBentrySTDinterwordspacing}{\spaceskip=0pt\relax}
\providecommand{\BIBentryALTinterwordstretchfactor}{4}
\providecommand{\BIBentryALTinterwordspacing}{\spaceskip=\fontdimen2\font plus
\BIBentryALTinterwordstretchfactor\fontdimen3\font minus
  \fontdimen4\font\relax}
\providecommand{\BIBforeignlanguage}[2]{{%
\expandafter\ifx\csname l@#1\endcsname\relax
\typeout{** WARNING: IEEEtran.bst: No hyphenation pattern has been}%
\typeout{** loaded for the language `#1'. Using the pattern for}%
\typeout{** the default language instead.}%
\else
\language=\csname l@#1\endcsname
\fi
#2}}
\providecommand{\BIBdecl}{\relax}
\BIBdecl

\bibitem{wang:gecco19}
R.~Wang, J.~Lehman, J.~Clune, and K.~O. Stanley, ``{POET}: Open-ended
  coevolution of environments and their optimized solutions,'' in
  \emph{Proceedings of the Genetic and Evolutionary Computation Conference},
  ser. GECCO '19.\hskip 1em plus 0.5em minus 0.4em\relax ACM, 2019, pp.
  142--151.

\bibitem{dharna2020cogeneration}
\BIBentryALTinterwordspacing
A.~Dharna, J.~Togelius, and L.~B. Soros, ``Co-generation of game levels and
  game-playing agents,'' \emph{Proceedings of the AAAI Conference on Artificial
  Intelligence and Interactive Digital Entertainment}, vol.~16, no.~1, pp.
  203--209, Oct. 2020. [Online]. Available:
  \url{https://ojs.aaai.org/index.php/AIIDE/article/view/7431}
\BIBentrySTDinterwordspacing

\bibitem{wang2020enhanced}
R.~Wang, J.~Lehman, A.~Rawal \emph{et~al.}, ``Enhanced {POET:} open-ended
  reinforcement learning through unbounded invention of learning challenges and
  their solutions,'' in \emph{Proceedings of the 37th International Conference
  on Machine Learning, {ICML} 2020, 13-18 July 2020, Virtual Event}, ser.
  Proceedings of Machine Learning Research, vol. 119.\hskip 1em plus 0.5em
  minus 0.4em\relax {PMLR}, 2020, pp. 9940--9951.

\bibitem{perez2014gvgai}
D.~Perez-Liebana, S.~Samothrakis, J.~Togelius \emph{et~al.}, ``The 2014 general
  video game playing competition,'' \emph{IEEE Transactions on Computational
  Intelligence and AI in Games}, vol.~8, no.~3, pp. 229--243, 2016.

\bibitem{MCTS2006}
\BIBentryALTinterwordspacing
L.~Kocsis and C.~Szepesvari, ``Bandit based monte-carlo planning.''\hskip 1em
  plus 0.5em minus 0.4em\relax Springer Berlin Heidelberg, 2006, pp. 282--293.
  [Online]. Available: \url{https://doi.org/10.1007/11871842_29}
\BIBentrySTDinterwordspacing

\bibitem{gmmTransfer2019}
H.~{Zuo}, J.~{Lu}, G.~{Zhang}, and F.~{Liu}, ``Fuzzy transfer learning using an
  infinite gaussian mixture model and active learning,'' \emph{IEEE
  Transactions on Fuzzy Systems}, vol.~27, no.~2, pp. 291--303, 2019.

\bibitem{bengioCL2009}
Y.~Bengio, J.~Louradour, R.~Collobert, and J.~Weston, ``Curriculum learning,''
  in \emph{Proceedings of the 26th Annual International Conference on Machine
  Learning}, ser. ICML '09.\hskip 1em plus 0.5em minus 0.4em\relax New York,
  NY, USA: Association for Computing Machinery, 2009, p. 41–48.

\bibitem{Lazaric2008batch}
A.~Lazaric, M.~Restelli, and A.~Bonarini, ``Transfer of samples in batch
  reinforcement learning,'' in \emph{Proceedings of the 25th International
  Conference on Machine Learning}, ser. ICML '08.\hskip 1em plus 0.5em minus
  0.4em\relax New York, NY, USA: Association for Computing Machinery, 2008, p.
  544–551.

\bibitem{Taylor09Negative}
M.~E. Taylor and P.~Stone, ``Transfer learning for reinforcement learning
  domains: A survey,'' \emph{J. Mach. Learn. Res.}, vol.~10, p. 1633–1685,
  Dec. 2009.

\bibitem{wangTranfserGP2020}
X.~Wang, Y.~Jin, S.~Schmitt, and M.~Olhofer, ``Transfer learning for gaussian
  process assisted evolutionary bi-objective optimization for objectives with
  different evaluation times,'' in \emph{Proceedings of the 2020 Genetic and
  Evolutionary Computation Conference}, ser. GECCO '20.\hskip 1em plus 0.5em
  minus 0.4em\relax New York, NY, USA: Association for Computing Machinery,
  2020, p. 587–594.

\bibitem{karpathy2014deep}
A.~{Karpathy} and L.~{Fei-Fei}, ``Deep visual-semantic alignments for
  generating image descriptions,'' in \emph{2015 IEEE Conference on Computer
  Vision and Pattern Recognition (CVPR)}, 2015, pp. 3128--3137.

\bibitem{finn2017modelagnostic}
C.~Finn, P.~Abbeel, and S.~Levine, ``Model-agnostic meta-learning for fast
  adaptation of deep networks,'' in \emph{Proceedings of the 34th International
  Conference on Machine Learning}, ser. Proceedings of Machine Learning
  Research, D.~Precup and Y.~W. Teh, Eds., vol.~70.\hskip 1em plus 0.5em minus
  0.4em\relax International Convention Centre, Sydney, Australia: PMLR, 06--11
  Aug 2017, pp. 1126--1135.

\bibitem{shelhamer2016loss}
\BIBentryALTinterwordspacing
E.~Shelhamer, P.~Mahmoudieh, M.~Argus, and T.~Darrell, ``Loss is its own
  reward: Self-supervision for reinforcement learning,'' \emph{CoRR}, vol.
  abs/1612.07307, 2016. [Online]. Available:
  \url{http://arxiv.org/abs/1612.07307}
\BIBentrySTDinterwordspacing

\bibitem{alex2019evolvability}
A.~Gajewski, J.~Clune, K.~O. Stanley, and J.~Lehman, ``Evolvability es:
  Scalable and direct optimization of evolvability,'' in \emph{Proceedings of
  the Genetic and Evolutionary Computation Conference}, ser. GECCO '19.\hskip
  1em plus 0.5em minus 0.4em\relax New York, NY, USA: Association for Computing
  Machinery, 2019, p. 107–115.

\bibitem{narvekarCurriculum}
S.~Narvekar, B.~Peng, M.~Leonetti, J.~Sinapov, M.~E. Taylor, and P.~Stone,
  ``Curriculum learning for reinforcement learning domains: A framework and
  survey,'' \emph{Journal of Machine Learning Research}, vol.~21, no. 181, pp.
  1--50, 2020.

\bibitem{MDP}
M.~L. Puterman, \emph{Markov Decision Processes: Discrete Stochastic Dynamic
  Programming}, 1st~ed.\hskip 1em plus 0.5em minus 0.4em\relax USA: John Wiley
  and Sons, Inc., 1994.

\bibitem{vizDoom2016}
\BIBentryALTinterwordspacing
D.~S. Chaplot, G.Lample, K.~M. Sathyendra, and R.~Salakhutdinov, ``Transfer
  deep reinforcement learning in 3d environments: An empirical study.'' in
  \emph{Proceedings of the 29th International Conference on Neural Information
  Processing Systems}, 2016. [Online]. Available:
  \url{https://www.cs.cmu.edu/~rsalakhu/papers/DeepRL_Transfer.pdf}
\BIBentrySTDinterwordspacing

\bibitem{nagab2018learning}
A.~Nagabandi, I.~Clavera, S.~Liu \emph{et~al.}, ``Learning to adapt in dynamic,
  real-world environments through meta-reinforcement learning,'' in
  \emph{Proceedings of the 35th International Conference on Machine Learning},
  2019.

\bibitem{Vinyals2019AStar}
O.~Vinyals, I.~Babuschkin, W.~M. Czarnecki \emph{et~al.}, ``Grandmaster level
  in {StarCraft} {II} using multi-agent reinforcement learning,''
  \emph{Nature}, vol. 575, no. 7782, pp. 350--354, Oct. 2019.

\bibitem{song2020rapidly}
X.~Song, Y.~Yang, K.~Choromanski \emph{et~al.}, ``Rapidly adaptable legged
  robots via evolutionary meta-learning,'' in \emph{2020 IEEE/RSJ International
  Conference on Intelligent Robots and Systems (IROS)}, 2020, pp. 3769--3776.

\bibitem{StarCraftTransfer2018}
K.~Shao, Y.~Zhu, and D.~Zhao, ``Starcraft micromanagement with reinforcement
  learning and curriculum transfer learning,'' \emph{IEEE Transactions on
  Emerging Topics in Computational Intelligence}, vol.~PP, 04 2018.

\bibitem{GoodfellowGAN}
I.~J. Goodfellow, J.~Pouget-Abadie, M.~Mirza \emph{et~al.}, ``Generative
  adversarial nets,'' in \emph{Proceedings of the 27th International Conference
  on Neural Information Processing Systems - Volume 2}, ser. NIPS’14.\hskip
  1em plus 0.5em minus 0.4em\relax Cambridge, MA, USA: MIT Press, 2014, p.
  2672–2680.

\bibitem{Arulkumaran_2019}
K.~Arulkumaran, A.~Cully, and J.~Togelius, ``Alphastar: An evolutionary
  computation perspective,'' \emph{Proceedings of the Genetic and Evolutionary
  Computation Conference Companion on - GECCO ’19}, 2019.

\bibitem{SilverGo2016}
D.~Silver, A.~Huang, C.~J. Maddison \emph{et~al.}, ``Mastering the game of go
  with deep neural networks and tree search,'' \emph{Nature}, vol. 529, pp.
  484--503, 2016.

\bibitem{openai2019dota}
\BIBentryALTinterwordspacing
OpenAI, C.~Berner, G.~Brockman \emph{et~al.}, ``Dota 2 with large scale deep
  reinforcement learning,'' 2019. [Online]. Available:
  \url{https://arxiv.org/abs/1912.06680}
\BIBentrySTDinterwordspacing

\bibitem{chellapilla:tec01}
K.~Chellapilla and D.~B. Fogel, ``Evolving an expert checkers playing program
  without using human expertise,'' \emph{IEEE Transactions on Evolutionary
  Computation}, vol.~5, no.~4, pp. 422--428, 2001.

\bibitem{Popovici2012}
E.~Popovici, A.~Bucci, R.~P. Wiegand, and E.~D. De~Jong, \emph{Coevolutionary
  Principles}.\hskip 1em plus 0.5em minus 0.4em\relax Berlin, Heidelberg:
  Springer Berlin Heidelberg, 2012, pp. 987--1033.

\bibitem{Kelly2018CoevoAtari}
\BIBentryALTinterwordspacing
S.~Kelly and M.~I. Heywood, ``{Emergent Solutions to High-Dimensional Multitask
  Reinforcement Learning},'' \emph{Evolutionary Computation}, vol.~26, no.~3,
  pp. 347--380, 09 2018. [Online]. Available:
  \url{https://doi.org/10.1162/evco\_a\_00232}
\BIBentrySTDinterwordspacing

\bibitem{Lanctot2017Unified}
M.~Lanctot, V.~Zambaldi, A.~Gruslys \emph{et~al.}, ``A unified game-theoretic
  approach to multiagent reinforcement learning,'' in \emph{Advances in Neural
  Information Processing Systems}, vol.~30, 2017.

\bibitem{guttenberg2018potential}
N.~Guttenberg, N.~Virgo, and A.~Penn, ``On the potential for open-endedness in
  neural networks,'' \emph{Artif. Life}, vol.~25, no.~2, p. 145–167, May
  2019.

\bibitem{LehmanMC}
J.~Lehman and K.~Stanley, ``Revising the evolutionary computation abstraction:
  Minimal criteria novelty search,'' in \emph{Proceedings of the 12th annual
  conference on Genetic and evolutionary computation}, 01 2010, pp. 103--110.

\bibitem{Soros2014IdentifyingNC}
L.~B. Soros and K.~O. Stanley, ``Identifying necessary conditions for
  open-ended evolution through the artificial life world of chromaria,''
  \emph{Artificial Life}, pp. 793--800, 2014.

\bibitem{soros18Thesis}
L.~Soros, ``Necessary conditions for open-ended evolution,'' Ph.D.
  dissertation, 08 2018.

\bibitem{perezliebana2018general}
D.~{Perez-Liebana}, J.~{Liu}, A.~{Khalifa}, R.~D. {Gaina}, J.~{Togelius}, and
  S.~M. {Lucas}, ``General video game ai: A multitrack framework for evaluating
  agents, games, and content generation algorithms,'' \emph{IEEE Transactions
  on Games}, vol.~11, no.~3, pp. 195--214, 2019.

\bibitem{schaul2014extensible}
T.~Schaul, ``An extensible description language for video games,'' \emph{IEEE
  Transactions on Computational Intelligence and AI in Games}, vol.~6, no.~4,
  pp. 325--331, 2014.

\bibitem{salimans2017:ESRL}
\BIBentryALTinterwordspacing
T.~Salimans, J.~Ho, X.~Chen, S.~Sidor, and I.~Sutskever, ``Evolution strategies
  as a scalable alternative to reinforcement learning,'' 2017. [Online].
  Available: \url{https://arxiv.org/abs/1703.03864}
\BIBentrySTDinterwordspacing

\bibitem{StornDE}
R.~Storn and K.~Price, ``Differential evolution: A simple and efficient
  adaptive scheme for global optimization over continuous spaces,''
  \emph{Journal of Global Optimization}, vol.~23, 01 1995.

\bibitem{Nelson2016MCTS}
M.~J. {Nelson}, ``Investigating vanilla mcts scaling on the gvg-ai game
  corpus,'' in \emph{2016 IEEE Conference on Computational Intelligence and
  Games (CIG)}, 2016.

\bibitem{lehman2011abandoning}
J.~Lehman and K.~O. Stanley, ``Abandoning objectives: Evolution through the
  search for novelty alone,'' \emph{Evolutionary computation}, vol.~19, no.~2,
  pp. 189--223, 2011.

\bibitem{huang2020action}
\BIBentryALTinterwordspacing
S.~Huang and S.~Ontañón, ``Action guidance: Getting the best of sparse
  rewards and shaped rewards for real-time strategy games,'' 2020. [Online].
  Available: \url{https://arxiv.org/abs/2010.03956}
\BIBentrySTDinterwordspacing

\bibitem{pathak2017curiositydriven}
D.~Pathak, P.~Agrawal, A.~A. Efros, and T.~Darrell, ``Curiosity-driven
  exploration by self-supervised prediction,'' ser. ICML'17.\hskip 1em plus
  0.5em minus 0.4em\relax JMLR.org, 2017, p. 2778–2787.

\bibitem{Silver2017}
D.~Silver, J.~Schrittwieser, K.~Simonyan \emph{et~al.}, ``Mastering the game of
  go without human knowledge,'' \emph{Nature}, vol. 550, no. 7676, pp.
  354--359, Oct. 2017.

\bibitem{LehmanLocalComp}
J.~Lehman and K.~O. Stanley, ``Evolving a diversity of virtual creatures
  through novelty search and local competition,'' in \emph{Proceedings of the
  13th Annual Conference on Genetic and Evolutionary Computation}, ser. GECCO
  '11.\hskip 1em plus 0.5em minus 0.4em\relax New York, NY, USA: Association
  for Computing Machinery, 2011, p. 211–218.

\bibitem{Stanley_evolvingneural}
K.~O. Stanley and R.~Miikkulainen, ``Evolving neural networks through
  augmenting topologies,'' \emph{Evolutionary Computation}, p. 2002.

\bibitem{mouret2015illuminating}
\BIBentryALTinterwordspacing
J.~Mouret and J.~Clune, ``Illuminating search spaces by mapping elites,''
  \emph{CoRR}, vol. abs/1504.04909, 2015. [Online]. Available:
  \url{http://arxiv.org/abs/1504.04909}
\BIBentrySTDinterwordspacing

\bibitem{Sidorov2013}
G.~Sidorov, F.~Velasquez, E.~Stamatatos, A.~Gelbukh, and
  L.~Chanona-Hern{\'{a}}ndez, ``Syntactic dependency-based n-grams as
  classification features,'' in \emph{Advances in Computational
  Intelligence}.\hskip 1em plus 0.5em minus 0.4em\relax Springer Berlin
  Heidelberg, 2013, pp. 1--11.

\bibitem{romesburg2004cluster}
C.~Romesburg, \emph{Cluster Analysis for Researchers}.\hskip 1em plus 0.5em
  minus 0.4em\relax Lulu.com, 2004.

\bibitem{gould_vrba_1982}
S.~J. Gould and E.~S. Vrba, ``Exaptation—a missing term in the science of
  form,'' \emph{Paleobiology}, vol.~8, no.~1, p. 4–15, 1982.

\bibitem{wang2018characterizing}
Z.~{Wang}, Z.~{Dai}, B.~{Poczos}, and J.~{Carbonell}, ``Characterizing and
  avoiding negative transfer,'' in \emph{2019 IEEE/CVF Conference on Computer
  Vision and Pattern Recognition (CVPR)}, 2019, pp. 11\,285--11\,294.

\bibitem{NgReward1999}
A.~Y. Ng, D.~Harada, and S.~J. Russell, ``Policy invariance under reward
  transformations: Theory and application to reward shaping,'' in
  \emph{Proceedings of the Sixteenth International Conference on Machine
  Learning}, ser. ICML '99.\hskip 1em plus 0.5em minus 0.4em\relax San
  Francisco, CA, USA: Morgan Kaufmann Publishers Inc., 1999, p. 278–287.

\bibitem{vinyals2017starcraft}
\BIBentryALTinterwordspacing
O.~Vinyals, T.~Ewalds, S.~Bartunov \emph{et~al.}, ``Starcraft ii: A new
  challenge for reinforcement learning,'' 2017. [Online]. Available:
  \url{https://arxiv.org/abs/1708.04782}
\BIBentrySTDinterwordspacing

\bibitem{grefenstette2019generalized}
\BIBentryALTinterwordspacing
E.~Grefenstette, B.~Amos, D.~Yarats \emph{et~al.}, ``Generalized inner loop
  meta-learning.'' \emph{CoRR}, vol. abs/1910.01727, 2019. [Online]. Available:
  \url{http://dblp.uni-trier.de/db/journals/corr/corr1910.html#abs-1910-01727}
\BIBentrySTDinterwordspacing

\bibitem{justesen2018illuminating}
\BIBentryALTinterwordspacing
N.~Justesen, R.~R. Torrado, P.~Bontrager, A.~Khalifa, J.~Togelius, and S.~Risi,
  ``Illuminating generalization in deep reinforcement learning through
  procedural level generation,'' 2018. [Online]. Available:
  \url{https://arxiv.org/abs/1806.10729/}
\BIBentrySTDinterwordspacing

\bibitem{ye2020rotation}
C.~Ye, A.~Khalifa, P.~Bontrager, and J.~Togelius, ``Rotation, translation, and
  cropping for zero-shot generalization,'' in \emph{IEEE Conference on Games,
  CoG 2020}, ser. IEEE Conference on Computatonal Intelligence and Games,
  CIG.\hskip 1em plus 0.5em minus 0.4em\relax IEEE Computer Society, Aug. 2020,
  pp. 57--64.

\bibitem{haarnoja2018soft}
T.~Haarnoja, A.~Zhou, P.~Abbeel, and S.~Levine, ``Soft actor-critic: Off-policy
  maximum entropy deep reinforcement learning with a stochastic actor,'' in
  \emph{Proceedings of the 35th International Conference on Machine Learning},
  ser. Proceedings of Machine Learning Research, J.~Dy and A.~Krause, Eds.,
  vol.~80.\hskip 1em plus 0.5em minus 0.4em\relax Stockholmsmässan, Stockholm
  Sweden: PMLR, 10--15 Jul 2018, pp. 1861--1870.

\bibitem{yu2019metaworld}
T.~Yu, D.~Quillen, Z.~He \emph{et~al.}, ``Meta-world: A benchmark and
  evaluation for multi-task and meta reinforcement learning,'' in
  \emph{Proceedings of the Conference on Robot Learning}, ser. Proceedings of
  Machine Learning Research, L.~P. Kaelbling, D.~Kragic, and K.~Sugiura, Eds.,
  vol. 100.\hskip 1em plus 0.5em minus 0.4em\relax PMLR, 30 Oct--01 Nov 2020,
  pp. 1094--1100.

\bibitem{Gaier2020Discovering}
A.~Gaier, A.~Asteroth, and J.-B. Mouret, ``Discovering representations for
  black-box optimization,'' \emph{Proceedings of the 2020 Genetic and
  Evolutionary Computation Conference}, Jun 2020.

\end{thebibliography}




\end{document}